\documentclass{article}

\usepackage[final]{neurips_2025}

\usepackage[utf8]{inputenc} 
\usepackage[T1]{fontenc}    
\usepackage{url}            
\usepackage{booktabs}       
\usepackage{amsfonts}       
\usepackage{nicefrac}       
\usepackage{microtype}      
\usepackage{xcolor}         

\usepackage{graphicx}
\usepackage{subfigure}
\usepackage{amsmath}
\usepackage{amsthm}
\usepackage{bm}
\usepackage{bbm}
\usepackage{wrapfig}

\usepackage{algorithm}
\usepackage{algorithmic}
\usepackage{graphicx}
\usepackage{textcomp}
\usepackage{subfigure} 
\usepackage{multirow}
\usepackage{multicol}
\usepackage{xcolor, colortbl}
\usepackage{bm}
\usepackage{bbm}
\usepackage[normalem]{ulem}
\usepackage{enumitem}
\usepackage{caption}
\usepackage{nicefrac}
\usepackage{makecell}
\newtheorem{theorem}{Theorem}
\newtheorem{lemma}{Lemma}

\def \mL{\mathcal L}

\newcommand{\bb}{\mathbf{b}}

\newcommand{\cD}{\mathcal{D}}

\newcommand{ \cH}{\mathcal{H}}

\usepackage{array}
\newcolumntype{L}[1]{>{\raggedright\let\newline\\\arraybackslash\hspace{0pt}}m{#1}}
\newcolumntype{C}[1]{>{\centering\let\newline  \\\arraybackslash\hspace{0pt}}m{#1}}
\newcolumntype{R}[1]{>{\raggedleft\let\newline \\\arraybackslash\hspace{0pt}}m{#1}}

\newcommand{\TheName}[0]{ConML}

\title{Learning to Learn with Contrastive Meta-Objective}

\author{%
	Shiguang Wu\textsuperscript{1 }, 
	Yaqing Wang\textsuperscript{2}$^*$,
	Yatao Bian\textsuperscript{3},
	Quanming Yao\textsuperscript{1,4}\thanks{Corresponding authors.}\\
	\textsuperscript{1}Department of Electronic Engineering, Tsinghua University\\
	\textsuperscript{2}Beijing Institute of Mathematical Sciences and Applications \\
	\textsuperscript{3} Department of Computer Science, National University of Singapore\\
	\textsuperscript{4}State Key laboratory of Space Network and Communications,
	Tsinghua University\\
	\texttt{wsg23@mails.tsinghua.edu.cn, wangyaqing@bimsa.cn,} \\
	\texttt{ybian@nus.edu.sg, qyaoaa@tsinghua.edu.cn}\\
}

\begin{document}
	
	\maketitle	
	
	\begin{abstract}
		Meta-learning enables learning systems to adapt quickly to new tasks, similar to humans.
		Different meta-learning approaches all work under/with the mini-batch episodic training framework. Such framework naturally gives the information about task identity, which can serve as additional supervision for meta-training to improve generalizability. We propose to exploit task identity as additional supervision in meta-training, inspired by the alignment and discrimination ability which is is intrinsic in human's fast learning.
		This is achieved by contrasting what meta-learners learn, i.e., model representations.
		The proposed \TheName{} is evaluating and optimizing the contrastive meta-objective under a problem- and learner-agnostic meta-training framework.
		We demonstrate that \TheName{} integrates seamlessly with existing meta-learners, 
		as well as in-context learning models, and brings significant boost in performance with small implementation cost.
	\end{abstract}
	
	\section{Introduction}
	
	Learning to learn, also known as meta-learning \citep{schmidhuber1987evolutionary,thrun1998learning}, 
	is a powerful paradigm designed to enable learning systems to adapt quickly to new tasks. 
	During the meta-training phase, a meta-learner simulates adaptation (learning) across a variety of relevant tasks to accumulate knowledge on how to learn effectively. In the meta-testing phase, this learned adaptation strategy is applied to unseen tasks. 
	The adaptation is typically accomplished by the meta-learner, which, given a set of task-specific training examples, generates a predictive model tailored to that task. 
	
	As the objective of meta-learning is to learn a meta-learner to generalize well to unseen tasks where a few labeled examples are given, 
	the most conventional objective in meta-training follows the natural idea "train as you test" \citep{vinyals2016matching} to minimize the validation loss, by splitting each task into a training set (support set) to which the meta-learner would be adapted to, and a validation set (query set) to evaluate the adapted model. Beyond "train as you test", people also have introduced regularization to the meta-training objective to improve generalizability, like supervision from stronger models \citep{wang2017learning,fei2021melr,ye2022few}, or injecting global information into each task \citep{wang2023robust}.
	All these works are under/with the same \textit{mini-batch episodic training} framework: 
	sampling a batch of tasks in each episode to obtain an episodic loss to minimize.
	
	The mini-batch episodic training framework is universal, and naturally gives the information about task identity, 
	which can serve as additional supervision for meta-training for generalizability.
	Inspired by the intrinsic property of human's fast learning ability: alignment and discrimination \citep{chen2012object,hummel2013object,christie2021learning},
	we hope \textbf{a meta-learner itself should be able to tell if different datasets are 
		from the same task or different tasks} by exploiting task identity. 
	A good learner possesses \textbf{alignment}  ability to align different partial views of a certain object, 
	which means they can integrate various aspects or perspectives of information to form a coherent understanding \citep{christian2021alignment}. 
	This means a meta-learner should learn similar models from different datasets of the same tasks even if the data are few or noisy, benefiting the meta-testing performance through being robust to the given labeled data.  \textbf{Discrimination} involves distinguishing between similar stimuli to respond appropriately only to decisive inputs. 
	This means a meta-learner should learn different models from different tasks even if some of their inputs are similar, benefiting meta-testing performance through generalization to diverse tasks.
	
	In this paper, we propose \TheName{}, 
	modifying the conventional mini-batch episodic meta-training with additional contrastive meta-objective to improve alignment and discrimination abilities of meta-learner.
	Similar to how contrastive learning contrasts unlabeled samples by identity, \TheName{} contrasts the outputs of the meta-learner based on task identity. Positive pairs consist of different subsets of the same task, while negative pairs come from different tasks, with the objective of minimizing inner-task distance (alignment) and maximizing inter-task distance (discrimination). 
	We design cheap and straightforward ways to obtain model representations for different types of meta-learners.

	\TheName{} distinguishes itself by being universal: it is \textbf{problem-agnostic}, as it is based-on mini-batch episodic training where task-identity are intrinsic information; and it is \textbf{learner-agnostic}, as we design easy-to-implement mapping functions from model to representations for different meta-learners. 
	Additionally, it is efficient in that it requires no additional data or retraining.
	Existing approaches have also
	leveraged task-level alignment or contrastiveness as additional supervision for improved meta-learning. 
	However, they are based on either problem-specific knowledge \citep{wang2017learning,ye2022few,wang2023robust}
	or learner-specific knowledge \citep{gondal2021function,fei2021melr}.
	Thus,
	they can be improved by exploiting the problem- and learner- agnostic task identity
	through incorporating with \TheName{}.
	Our contributions are:
	\begin{itemize}[leftmargin=*]
		\item 
		We propose to exploit task-identity as additional supervision in meta-training by emulating human cognitive alignment and discrimination abilities.
		
		\item 
		We extend contrastive learning from the representation space in unsupervised learning to the model space in meta-learning, by designing mapping functions from models to representations for various types of meta-learners.
		
		\item 
		We empirically show the proposed \TheName{} universally improves the performance of various meta-learning algorithms from different categories with small implementation cost.
		Furthermore, we show that \TheName{} can also improve in-context learning (ICL) as its training also follows the paradigm of learning to learn.
	\end{itemize}
	
	\begin{figure}[t]
		\centering
		\includegraphics[width=0.90\textwidth]{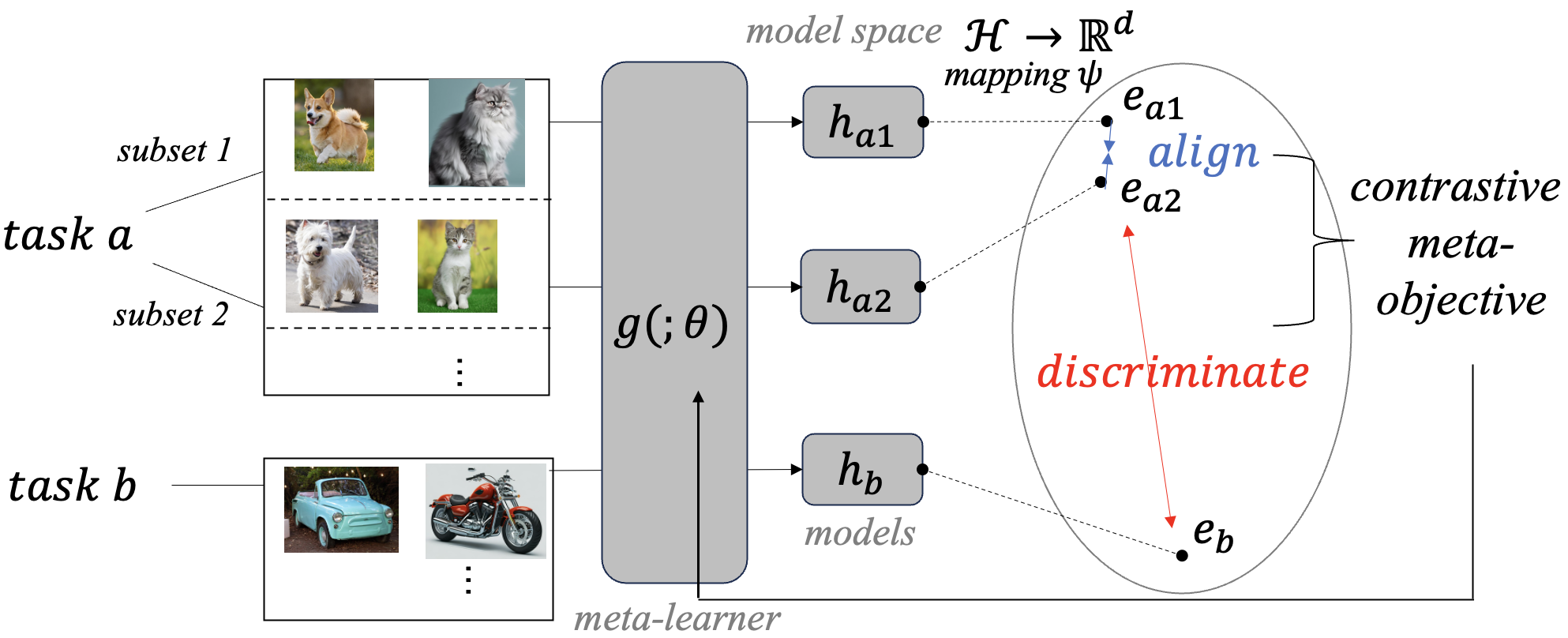}
		\label{fig:illus}
		\caption{\TheName{} is performing contrastive learning in model space, 
			to make the meta-learner itself able to align information from the same task (alignment) 
			while discriminate different tasks to improve generalizability (discrimination).}
	\end{figure}

	\section{Preliminaries: Learning to Learn}
	\label{sec:rw}
	
	Learning to learn, which is also known as meta-learning, 
	focuses on improving the learning algorithm itself \citep{schmidhuber1987evolutionary}. 
	We focus on the most general meta-learning setting.
	Formally, let $g(;\theta)$ be a meta-learner that maps a dataset $\cD$ to a model $h$, i.e., $h = g(\cD; \theta)$. Let $\mL(\cD;h)$ represent the loss when evaluating a model $h$ on a dataset $\cD$ using a loss function $\ell(y, \hat{y})$ (e.g., cross-entropy or mean squared error). 
	Given a distribution of tasks $p(\tau)$ for meta-training, where each task $\tau$ corresponds to a dataset $\cD_\tau = \{(x_{\tau,i}, y_{\tau,i})\}_{i=1}^m$, the objective of meta-learning is to train $g(;\theta)$ to generalize well to unseen task $\tau'$ sampled from $p(\tau')$. 
	During meta-testing, give an unseen task $\tau'$ with labeled dataset $\cD^{\text{tr}}_{\tau'}$ (training set or support set) to $g(;\theta)$ to generate $h$, which is tested by another set from the same task $\cD^{\text{val}}_{\tau'}$ (validation set or query set), i.e., evaluated by $\mL(\cD^{\text{val}}_{\tau'}; g(\cD^{\text{tr}}_{\tau'}; \theta))$.

	In meta-training, the meta-learner $g(;\theta)$ is optimized through a series of episodes, each consisting of a batch $\bb$ of $B$ tasks, and obtains an episodic loss $\mL_e$ to minimize. The form of $\mL_e$ can be various, while we take the most typical validation loss as example to illustrate our method. Splitting each $\cD_\tau$ into a training set $\cD^{\text{tr}}_\tau = \{(x_{\tau,i}, y_{\tau,i})\}_{i=1}^n$ and a validation set $\cD^{\text{val}}_\tau = \{(x_{\tau,i}, y_{\tau,i})\}_{i=n+1}^m$, the meta-training objective is minimizing $\mathbb{E}_{\tau \sim p(\tau)} \mL(\cD^{\text{val}}_{\tau}; g(\cD^{\text{tr}}_{\tau}; \theta))$. The mini-batch episodic training with validation loss is outlined in Algorithm \ref{alg:episodic-training}.
	Note that \TheName{} relies on the mini-batch episodic framework, which is general.  \TheName{} does not rely on how the specific meta-learner measures $\mL_e$ inside each episode. Here we take the representative validation loss as example and will discuss other forms in Section~\ref{sec:relate}.
	
	\begin{wrapfigure}{R}{0.45\textwidth}
		\vspace{-20px}
		\begin{minipage}{0.45\textwidth}
			\begin{algorithm}[H]
				\small
				\caption{Mini-Batch Episodic Training (with Validation Loss).}
				\label{alg:episodic-training}
				\begin{algorithmic}
					\WHILE {Not converged}
					\STATE Sample a batch of tasks $\bm{b}\sim p^B(\tau)$.
					\FOR { All $\tau\in \bm{b}$}
					\STATE Get task-specific model $h_{\tau}=g(\cD^{\text{tr}}_{\tau};\theta)$;
					\STATE Get validation loss $\mL(\cD^{\text{val}}_{\tau};h_{\tau})$;
					\ENDFOR
					\STATE Get episodic loss \\$\mL_e = \frac{1}{B} \sum_{\tau\in\bm{b}}\mL(\cD^{\text{val}}_{\tau};g(\cD^{\text{tr}}_{\tau};\theta))$;
					\STATE Update $\theta$ by $\theta\leftarrow\theta-\nabla_\theta \mL_e$.
					\ENDWHILE
				\end{algorithmic}
			\end{algorithm}
		\end{minipage}
		\vspace{-10px}
	\end{wrapfigure}

	Different meta-learners implement their own specific functions within $g(;\theta)$. 
	Popular meta-learning approaches can be broadly categorized into the following types \citep{bronskill2021memory}: (i) Optimization-based approaches \citep{andrychowicz2016learning,finn2017model,nichol2018first}, which focus on learning better optimization strategies for adapting to new tasks; (ii) Metric-based approaches \citep{vinyals2016matching,snell2017prototypical,sung2018learning}, which leverage learned similarity metrics; and (iii) Amortization-based approaches \citep{garnelo2018conditional,requeima2019fast,bateni2020improved}, which aim to learn a shared representation across tasks, amortizing the adaptation process by using neural networks to directly infer task-specific parameters from the training set; (iv) Furthermore, the emerging ICL ability in large language models (LLMs) can also be viewed as the consequence of meta-learning \cite{garg2022can,akyurek2023learning}, and ICL model is meta-learner with minimal inductive bias \citep{wu2025context}, so we will also use meta-learner $g$ to express the function of ICL. Details reformulating ICL model as meta-learner to incorporate \TheName{} with are in Section~\ref{sec:species}. 
	
	\section{Meta-Learning with \TheName{}}
	\label{sec:3step}
	
	Now, we introduce our \TheName{} 
	which equips meta-learners with the desired alignment and discrimination ability via task-level contrastive learning.

	\subsection{A General Framework}
	\label{sec:learn-general}
	
	To enhance the alignment and discrimination abilities of meta-learning, we draw inspiration from Contrastive Learning (CL) \citep{oord2018representation,chen2020simple,wang2020understanding}. CL focuses on learning representations that are invariant to irrelevant details while preserving essential information. This is achieved by maximizing alignment and discrimination (uniformity) in the representation space \citep{wang2020understanding}. While most existing studies focus on sample-wise contrastive learning in the representation space via unsupervised learning \citep{hjelm2018learning,bachman2019learning,tian2020contrastive,chen2020simple}, we extend CL to the model space in meta-learning.  
	
	Specifically, we introduce contrastive meta-objective by tasks-level CL in the model space, where alignment is achieved by minimizing the inner-task distance (i.e., the distance between models trained on different subsets of the same task), and discrimination is achieved by maximizing the inter-task distance (i.e., the distance between models from different tasks). Such alignment and discrimination together form the contrastive meta-objective to optimize a meta-learner.
	The detailed procedures of \TheName{} are introduced below.

	\textit{Obtaining Model Representation.~}
	To train the meta-learner $g$, 
	the inner-task distance $d^{\text{in}}$ and inter-task distance $d^{\text{out}}$ are measured in the 
	output space of $g$, also referred to as the model space $\cH$. 
	A practical approach is to represent the model 
	$h = g(\cD; \theta) \in \cH$ as a fixed-length vector $\bm{e} \in \mathbb{R}^d$, and then compute the distances using an explicit distance function $\phi(\cdot, \cdot)$ (e.g., cosine distance). 
	To form a learner-agnostic framework, we introduce a projection function $\psi: \cH \rightarrow \mathbb{R}^d$ to obtain the model representations $\bm{e} = \psi(h)$.
	The details of $\cH$ and $\psi$ will be specified in Section \ref{sec:species}. 

	\textit{Obtaining Inner-Task Distance.~}
	Alignment is achieved by minimizing inner-task distance.
	During meta-training, 
	the combined dataset 
	$\cD^{\text{tr}}_\tau\cup\cD^{\text{val}}_\tau$ contains all the available information about task $\tau$. 
	The meta-learner is expected to produce similar models when trained on any subset $\kappa$ of this dataset. 
	Moreover, models trained on subsets should resemble the model learned from the full dataset  $\cD^{\text{tr}}_\tau\cup\cD^{\text{val}}_\tau$.
	For $\forall \kappa\subseteq\cD^{\text{tr}}_\tau\cup\cD^{\text{val}}_\tau$,  we expect
	$\bm{e}^{\kappa}_{\tau}=\bm{e}^*_{\tau}$,
	where $\bm{e}^{\kappa}_{\tau}=  \psi(g(\kappa;\theta)),\bm{e}^*_{\tau}=  \psi(g(\cD^{\text{tr}}_\tau\cup\cD^{\text{val}}_\tau;\theta))$. 
	The inner-task distance $d^{\text{in}}_\tau$ for each task $\tau$ is computed as:
	\begin{align}
		\label{eq:ind}
		d^{\text{in}}_{\tau}
		=  ({1}/{K})\cdot\sum\nolimits_{k=1}^{K} \phi(\bm{e}^{\kappa_k}_{\tau},\bm{e}^*_{\tau}),
		\quad
		\text{s.t.\;\;}
		\kappa_k\sim\pi_\kappa(\cD^{\text{tr}}_\tau\cup\cD^{\text{val}}_\tau),
	\end{align}
	where $\{\kappa_k\}_{k=1}^K$ are $K$ subsets sampled from $\cD^{\text{tr}}_\tau\cup\cD^{\text{val}}_\tau$ using a specific sampling strategy $\pi_\kappa$. 
	In each episode, given a batch $\bb$ of task  containing $B$ tasks, the overall inner-task distance is averaged as $d^{\text{in}}=\frac{1}{B}\sum_{\tau\in\bm{b}}d^{\text{in}}_\tau$.

	\textit{Obtaining Inter-Task Distance.~}
	Discrimination is achieved by maximizing inter-task distance.
	Since the goal of meta-learning is to improve performance on unseen tasks, it is crucial for the meta-learner $g$ to generalize well across diverse tasks. Given the natural assumption that different tasks require distinct task-specific models, it is essential that $g$ can learn to differentiate between tasks—i.e., possess strong discrimination capabilities. 
	To enhance task-level generalization, we define the inter-task distance $d^{\text{out}}$, which should be maximized to encourage $g$ to learn distinct models for different tasks. Specifically, for any two tasks $\tau \neq \tau'$ during meta-training, 
	we maximize the distance between their respective representations,  $\bm{e}^*_{\tau}$ and $\bm{e}^*_{\tau'}$. 
	To make this practical within the mini-batch episodic training paradigm, we compute $d^{\text{out}}$ across a batch of tasks sampled in each episode: 
	\begin{align}
		\label{eq:xd}
		d^{\text{out}} 
		= 
		({1}/{B(B-1)})\cdot
		\sum\nolimits_{\tau\in \bm{b}}
		\sum\nolimits_{{\tau'}\in \bm{b}\backslash\tau}\phi(\bm{e}^*_{\tau},\bm{e}^*_{\tau'}).
	\end{align}

	\begin{wrapfigure}{R}{0.56\textwidth}
		\vspace{-22px}
		\begin{minipage}{0.56\textwidth}
			\begin{algorithm}[H]
				\small
				\caption{Meta-Training with \TheName{} (with Validation Loss).} 
				\label{alg:general}
				\begin{algorithmic}
					\WHILE {Not converged}
					\STATE Sample a batch of tasks $\bm{b}\sim p^B(\tau)$.
					\FOR { All $\tau\in \bm{b}$}
					\STATE  \dag{}Sample $\kappa_k$ from $\pi_\kappa(\cD^{\text{tr}}_\tau\cup\cD^{\text{val}}_\tau)$ for $k\in\{1\cdots K\}$;
					\STATE  \dag{}Get model representation $\bm{e}^{\kappa_k}_{\tau}=  \psi(g(\kappa_k;\theta))$;
					\STATE  \dag{}Get model representation $\bm{e}^*_{\tau}\!=\! \psi(g(\cD^{\text{tr}}_\tau\cup\cD^{\text{val}}_\tau;\theta))$;
					\STATE  \dag{}Get inner-task distance $d^{\text{in}}_{\tau}$ by \eqref{eq:ind};
					\STATE Get task-specific model $h_{\tau}=g(\cD^{\text{tr}}_{\tau};\theta)$;
					\STATE Get validation loss $\mL(\cD^{\text{val}}_{\tau};h_{\tau})$;
					\ENDFOR
					\STATE  \dag{}Get $d^{\text{in}}=\frac{1}{B}\sum_{\tau\in\bm{b}}d^{\text{in}}_\tau$ and $d^{\text{out}}$ by \eqref{eq:xd};
					\STATE Get loss $\mL_{\text{\TheName}}$ by \eqref{eq:obj};
					\STATE Update $\theta$ by $\theta\leftarrow\theta-\nabla_\theta \mL$.
					\ENDWHILE
				\end{algorithmic}
			\end{algorithm}
			\vspace{-10px}
			{\small 
				"\dag{}" indicates additional steps introduced by \TheName{} to Algorithm~\ref{alg:episodic-training}.}
		\end{minipage}
	\end{wrapfigure}

	\textit{Training Procedure.}
	\TheName{} optimizes
	the combination of the original episodic loss $\mL_e$ and contrastive meta-objective $\mL_c=d^{\text{in}} - d^{\text{out}}$:
	\begin{align}\label{eq:obj}
		\mL_{\text{\TheName}}=\mL_e+\lambda \mL_c
	\end{align}
	The meta-training procedure with \TheName{} is in Algorithm \ref{alg:general}. 
	Note \TheName{} is agnostic to the form of $\mL_e$ so here we take the typical validation loss as example.
	Compared to Algorithm~\ref{alg:episodic-training}, \TheName{} introduces additional computations for $\psi(g(\cD; \theta))$ a total of $K+1$ times per episode. However, $\psi$ is implemented as a lightweight function (e.g., extracting model weights), and $g(\cD; \theta)$ is already part of the standard episodic training process, with multiple evaluations of $g(\cD; \theta)$ being parallelizable. 
	As a result, \TheName{} incurs only a 
	little extra cost in computation
	(detailed analysis is in Appendix~\ref{app:complex}).
	
	\subsection{Provable Benefits for Generalization}\label{sec:theory}
	Here, we provide another perspective to understand how \TheName{} helps meta-learning.
	It is provable that a meta-learner which minimizes $\mL_c$ has lower generalization error upper-bound than any other meta-learners. 
	This means \TheName{} serves as a `safeguard' for the worst case of error due to finite samples in $\cD^{\text{val}}_\tau$ in meta-testing, that can be plugged-in any meta-learners.
	
	Following \cite{maurer2016benefit}, 
	the excess risk of a meta-learner $g(;\theta)$ is defined as:
	\begin{align*}
		\Delta\epsilon_{p(\tau)}(\theta)
		\! = \! {E_{\tau \sim p(\tau)}}   {E_{D_{\tau}^{tr}\sim \tau^n}}{E_{(x,y)\sim \tau}}
		\ell\big(g(D_{\tau}^{tr};\theta)(x),y \big)
		\!\! - \!\!
		\min_{\theta}
		\!
		{E_{\tau\sim p(\tau)}}
		\Big[\!
		\min_{h\in H_{g(\theta)}}{ 
			\!\!\!\! E_{(x,y)\sim \tau}}
		\ell\big( h(x),y \big)
		\Big],
		\!\!
	\end{align*}
	where $H_{g(\theta)}$ is the hypothesis class of $h$ given $g(;\theta)$. The value of $\Delta\epsilon_{{p(\tau)}}(\theta)>0$, means the difference between expectation of validation loss between $g(;\theta)$ given finite $n$ examples per task, 
	and the best we can find given $g$ and $p(\tau)$. 
	First, we can find an upper bound $U_{p(\tau)}(\theta)$ for  
	$\Delta\epsilon_{p(\tau)}(\theta)$. 
	
	\begin{lemma}
		\label{lem:1}
		Denote $ U_{p(\tau)}(\theta)=C_1\sqrt{\sup_{||v||\leq 1}{E_{\tau\sim p(\tau)}} {E_{(x,y)\sim \tau}}[\langle v,g(\{(x,y)\};\theta)\rangle^2]}+C_2$. There exists positive constants $C_1,C_2$ not related with $\theta$, satisfying
		$\forall \theta$, $\Delta\epsilon_{p(\tau)}(\theta) \leq U_{p(\tau)}(\theta)$.
	\end{lemma}

	Given contrastive meta-objective $\mL_c$ as defined above, with mild assumptions and choice of $\phi$ we then have the following theorem:
	\begin{theorem}
		\label{theory1}
		$\forall p(\tau),~U_{p(\tau)}(\theta^*_{\mL_c})=\min_{\theta}U_{p(\tau)}(\theta)$, where $\theta^*_{\mL_c}=\arg\min_{\theta}\mL_c(g(;\theta),p(\tau))$.
	\end{theorem}
	This means the contrastive meta-objective can exactly serve as a surrogate objective of the worst-case meta-testing performance, as described in the above theorem. Note that this holds for any $p(\tau)$ and $g$, which indicates the problem- and learner-agnostic benefit of \TheName{}.
	The proof is in Appendix~\ref{app:theory}.

	\subsection{Integrating with Typical Meta-Learners}
	\label{sec:species}
	
	\TheName{} is universally applicable to enhance  meta-learning algorithm that follows episodic training. It does not depend on a specific form of $g$ or $\mL_e$ and can be used alongside other forms of task-level information. 
	Next, we provide the specifications of $\cH$ and $\psi(g(\cD, \theta))$ to obtain model representations for implementing \TheName{}. 
	We illustrate examples across different categories of meta-learning algorithms, including optimization-based, metric-based, amortization-based and ICL. 
	They are summarized in Table~\ref{tab:specify}. 
	Appendix~\ref{app:algs} provides
	the detailed procedures for integrating \TheName{} with various meta-learning algorithms.

	\begin{table*}[t]
		\vspace{-5pt}
		\centering
		\small
		\setlength\tabcolsep{5pt}
		\caption{Specifications  of integrating \TheName{} with different meta-learners.}
		\vspace{-5px}
		\begin{tabular}{c|c|c|c}
			\toprule
			Category &Examples&Meta-learner $g(\cD;\theta)$&Model representation $\psi(g(\cD;\theta))$\\\midrule
			\makecell{Optimization\\-based}&	\makecell{MAML,\\Reptile}&	\makecell{Update model weights \\$\theta-\nabla_\theta\mL(\cD;h(;\theta))$}&\makecell{Updated model weight\\ $\theta-\nabla_\theta\mL(\cD;h(;\theta))$}\\\midrule
			\makecell{Metric\\-based}&\makecell{ProtoNet,\\ MatchNet }&\makecell{Build classifier with\\$\{(\{f(x_i;\theta)\}_{x_i\in\cD_j},\text{label }j)\}_{j=1}^N$}&\makecell{Concatenate\\$[\frac{1}{|\cD_j|}\sum_{x_i\in\cD_j}f(x_i;\theta)]_{j=1}^N$}\\\midrule
			\makecell{Amortization\\-based}&\makecell{CNPs,\\CNAPs }&\makecell{Map $\cD$ to model weights\\ by $H(\cD;\theta)$}&Output of hypernetwork $H(\cD;\theta)$	\\\midrule
			\makecell{In-context\\learning}&\makecell{In-context\\learning}&\makecell{Task-specific prediction for $x$ is given by\\sequential model $g([\vec\cD,x];\theta)$}&\makecell{$g([\vec\cD,u];\theta),$\\where $u$ is dummy input}\\\bottomrule
		\end{tabular}
		\label{tab:specify}
		\vspace{-10pt}
	\end{table*}
	
	\textit{With Optimization-Based.~}
	The representative algorithm of optimization-based meta-learning is
	MAML, which meta-learns an initialization from where gradient steps are taken to learn task-specific models, i.e., $g(\cD;\theta)=h(;{\theta-\nabla_\theta\mL(\cD;h(;\theta))})$. 
	Since MAML directly generates the model weights, we use these weights as model representation. 
	Specifically, the representation of the model learned by $g$ given a dataset $\cD$ is: 
	$
	\psi(g(\cD;\theta))=\theta-\nabla_\theta\mL(\cD;h(;\theta)),
	$
	certain optimization-based meta-learning algorithms, such as FOMAML \citep{finn2017model} and Reptile \citep{nichol2018first}, use first-order approximations of MAML and do not strictly follow Algorithm \ref{alg:episodic-training} to minimize validation loss. Nonetheless, \TheName{} can still be incorporated into these algorithms as long as they adhere to the episodic training framework.
	
	\textit{With Metric-Based.~}
	Metric-based algorithms are well-suited for classification tasks.
	Given a dataset $\cD$ for an $N$-way classification task, these algorithms classify based on the distances between input samples $\{\{f(x_i;\theta)\}_{x_i \in \cD_j}\}_{j=1}^N$ and their corresponding labels, where $f(;\theta)$ is a meta-learned encoder and $\cD_j$ represents the set of inputs for class $j$. We represent this metric-based classifier by concatenating the mean embeddings of each class in a label-aware order. 
	For example, ProtoNet  \citep{snell2017prototypical} computes the prototype $\bm{c}_j$, which is the mean embedding of samples in each class:
	$\bm{c}_j=\frac{1}{|\cD_j|}\sum_{(x_i,y_i)\in\cD_j}f(x_i;\theta)$. 
	The classifier $h_{\tau}$ then makes predictions as 
	$p(y=j\mid x) = \exp(-d(f(x;\theta),\bm{c}_j)) / \sum_{j'} \exp(-d(f(x;\theta),\bm{c}_{j'}))$.  
	Since the outcome model $h_{\tau}$ depends on $\cD$ through $\{\bm{c}_j\}_ {j=1}^N$ and their corresponding labels, 
	the representation is specified as
	$
	\psi(g(\cD;\theta))=[\bm{c}_1|\bm{c}_2|\cdots|\bm{c}_N]
	$, 
	where $[\cdot|\cdot]$ denotes concatenation.

	\textit{With Amortization-Based.~}
	Amortization-based approaches meta-learns a hypernetwork $H(;\theta)$ that aggregates information from $\cD$ to task-specific parameter $\alpha$, which serves as the weights for the main-network $h$, resulting in a task-specific model $h(;\alpha)$. 
	For example, Simple CNAPS  \citep{bateni2020improved} uses a hypernetwork to generate a small set of task-specific parameters that  perform feature-wise linear modulation (FiLM) on the convolution channels of the main-network.
	In \TheName{}, we represent the task-specific model $h(;\alpha)$ using the task-specific parameters $\alpha$, i.e., the output of the hypernetwork $H(;\theta)$:
	$
	\psi(g(\cD;\theta))=  H(\cD;\theta)
	$.
	
	\textit{With In-Context Learning (ICL).~}
	An ICL model makes task-specific prediction by $g([\vec\cD,x];\theta)$, 
	where $g$ is a sequential model and $\vec{\cD}$ is the sequentialized $\cD$ (prompt), $[x_{1},y_{1},\cdots,x_{m},y_{m}]$. 
	The details are in Appendix~\ref{sec:icl}.
	Note that ICL does not specify an explicit output model $h(x)=g(\cD;\theta)(x)$; instead, this procedure exists only implicitly through the feeding-forward of the sequence model. 
	Thus, obtaining the
	representation $\psi(g(\cD;\theta))$ by explicit model weights of $h$ is not feasible for ICL.
	To represent what $g$ learns from $\cD$, we design to incorporate $\vec{\cD}$ with a dummy input $u$, 
	which functions as a probe and its corresponding output can be readout as representation:
	\begin{align}\label{eq:repre-icl}
		\psi(g(\cD;\theta))=g([\vec\cD,u];\theta),
	\end{align}
	where $u$ is constrained to be in the same shape as $x$, and has consistent value in an episode.
	For example, for training a ICL model on linear regression tasks we can choose $u=\bm{1}$, and in pretraining of LLM we can choose $u=$"\textit{what is this task?}".
	The complete algorithm of \TheName{} for training an ICL model is in Appendix~\ref{app:algs}.
	
	\section{Empirical Studies}\label{sec:exp}
	
	We provide empirical studies to understand the effect of \TheName{} on synthetic data, 
	which shows that learning to learn with \TheName{} brings generalizable alignment and discrimination abilities.
	Code is avaliable  at \url{https://github.com/LARS-research/ConML}.
	
	\begin{table}[b]
		\centering
		\scriptsize
		\vspace{-15px}
		\caption{Meta-testing accuracy (\%) on \textit{mini}ImageNet and \textit{tiered}ImageNet.}
		\begin{tabular}{c|c|c|c|c|c|c}
			\toprule
			\multirow{2}{*}{Category} & \multirow{2}{*}{Algorithm} &  \multirow{2}{*}{\makecell{Objective} }& \multicolumn{2}{c}{\textit{mini}ImageNet }
			& \multicolumn{2}{c}{\textit{tiered}ImageNet }\\ 
			&&&5-way 1-shot &5-way 5-shot &
			5-way 1-shot & 5-way 5-shot\\ \midrule
			\multirow{6}{*}{\makecell{Optimization-\\Based} }& \multirow{2}{*}{MAML}  & - & $48.75\pm1.25$ & $64.50\pm1.02$ & $51.39\pm1.31$ & $68.25\pm0.98$ \\
			& &  w/ ConML&$\bm{56.25\pm0.94}$  & $\bm{67.37\pm0.97}$ & $\bm{58.75\pm1.45}$ & $\bm{72.94\pm0.98}$\\\cmidrule{2-7}
			& \multirow{2}{*}{FOMAML}   &  - & $48.12\pm1.40$ &  $63.86\pm0.95$ & $51.44\pm1.51$ & $68.32\pm0.95$\\
			& & w/ ConML &$\bm{57.64\pm1.29}$ & $\bm{68.50\pm0.78}$& $\bm{58.21\pm1.22}$ & $\bm{73.26\pm0.78}$ \\ \cmidrule{2-7}
			& \multirow{2}{*}{Reptile}  &  - & $49.21\pm0.60$ & $64.31\pm0.97$ & $47.88\pm1.62$ & $65.10\pm1.13$\\
			& &  w/ ConML & $\bm{52.82\pm1.06}$ & $\bm{67.04\pm0.81}$ & $\bm{55.01\pm1.28}$  & $\bm{70.15\pm1.00}$ \\ \midrule
			\multirow{4}{*}{\makecell{Metric-\\Based} } & \multirow{2}{*}{MatchNet}  &  - & $43.92\pm1.03$ & $56.26\pm0.90$ & $48.74\pm1.06$ & $61.30\pm0.94$  \\
			&  & w/ ConML & $\bm{48.75\pm0.88}$  & $\bm{62.04\pm0.89}$& $\bm{53.29\pm1.05}$ & $\bm{67.86\pm0.77}$   \\\cmidrule{2-7}
			& \multirow{2}{*}{ProtoNet}   &  - & $48.90\pm0.84$ & $65.69\pm0.96$ & $52.50\pm0.96$ & $71.03\pm0.74$  \\
			&  & w/ ConML & $\bm{51.03\pm0.91}$ & $\bm{67.35\pm0.72}$& $\bm{54.62\pm0.79}$& $\bm{73.78\pm0.75}$  \\ \midrule
			\multirow{2}{*}{\makecell{Amortization-\\Based} }& \multirow{2}{*}{SCNAPs}  &  - & $53.14\pm0.88$ & $70.43\pm0.76$ & $62.88\pm1.04$ & $79.82\pm0.87$ \\
			&  &  w/ ConML &$\bm{55.73\pm0.86}$ & $\bm{71.70\pm0.71}$& $\bm{65.06\pm0.95}$ & $\bm{81.79\pm0.80}$   \\\midrule
			\multirow{2}{*}{\makecell{In-Context\\Learning} }& \multirow{2}{*}{CAML}  &  - & $96.15\pm0.10$ & $98.57\pm0.08$ & $95.41\pm0.10$ & $98.06\pm0.10$ \\
			&  &  w/ ConML &$\bm{97.03\pm0.10}$ & $\bm{98.92\pm0.08}$& $\bm{96.56\pm0.09}$ & $\bm{98.23\pm0.05}$   \\\midrule
			\multirow{4}{*}{\makecell{Other\\Objective} }& \multirow{2}{*}{MELR} 
			&  - & $51.33 \pm 0.73$ & $68.16 \pm 0.59$ & $54.96\pm 0.89$ & $72.51 \pm 0.81$\\
			& &  w/ ConML& $\bm{53.56\pm1.02}$ & $\bm{70.04\pm0.95}$&$\bm{57.06\pm0.90}$ & $\bm{74.21\pm0.78}$  \\ \cmidrule{2-7}
			& \multirow{2}{*}{LastShot}  
			&  - & $64.80\pm 0.20$ & $81.65\pm 0.14$& $69.37\pm0.23$ & $85.36 \pm0.16$  \\
			& &  w/ ConML & $\bm{66.24\pm0.72}$ & $\bm{83.29\pm0.45}$ & $\bm{71.82\pm0.70}$ & $\bm{87.05\pm0.49}$  \\ 
			\bottomrule
		\end{tabular}
		\label{tab:mini}
		\vspace{-10pt}
	\end{table}
	
	\subsection{Few-Shot Image Classification Performance}
	\label{sec:exp-fscv}

	To show \TheName{} brings learner-agnostic improvement, we integrate \TheName{} into various meta-learners  
	and evaluate the meta-learning performance on few-shot image classification problem follow existing works \citep{vinyals2016matching,finn2017model,bateni2020improved}.
	We use two few-shot image classification benchmarks: miniImageNet \citep{vinyals2016matching} and tieredImageNet \citep{ren2018meta}, evaluating on 5-way 1-shot and 5-way 5-shot tasks.

	We consider representative meta-learning algorithms from different categories, including optimization-based: {MAML} \citep{finn2017model}, {FOMAML} \citep{finn2017model}, {Reptile} \citep{nichol2018first};
	metric-based: {MatchNet} \citep{vinyals2016matching}, {ProtoNet} \citep{snell2017prototypical}; amortization-based: {SCNAPs} (Simple CNAPS) \citep{bateni2020improved}; and the state-of-the-art ICL-based few-shot learner: {CAML} \citep{fifty2024context}. Note that for CAML, \TheName{} only effect the meta-training of the ICL mode, not the pretraining of Vit feature extractor.
	We also incorporate \TheName{} with meta-learners with improved meta-training objective as discussed in Section~\ref{sec:relate}, including: {MELR} \citep{fei2021melr} and {LastShot} \citep{ye2022few}.
	We evaluate the meta-learning performance of each algorithm in its original form (w/o \TheName{}) and after incorporating \TheName{} into the training process (w/ \TheName{}). The implementation of \TheName{} follows the general procedure described in Algorithm \ref{alg:general} and the specification for corresponding category in Section \ref{sec:species}.

	Table~\ref{tab:mini} shows the results on \textit{mini}ImageNet and \textit{tiered}ImageNet respectively. 
	We uses a common configuration for \TheName{}'s hyperparameter for 
	all meta-learners: 
	task batch size $B=32$, inner-task sampling $K=1$, and $\pi_\kappa(\cD^{\text{tr}}_\tau\cup\cD^{\text{val}}_\tau)=\cD^{\text{tr}}_\tau$, $\phi(a,b)=1-\nicefrac{a\cdot b}{\Vert a\Vert\Vert b\Vert}$ and $\lambda=0.1$. 
	Other hyperparameters related to model architecture and training procedure remain consistent with the original meta-learners'. 
	This demonstrates boosted performance can be brought even without specific hyperparameter tuning for different meta-learners.
	The performance improvement
	demonstrates that \TheName{} offers universal improvements across different meta-learning algorithms.
	Note that performance between different algorithms are not comparable. We also show \TheName{}'s consistent benefit on different sizes of backbones in Appendix~\ref{app:backbone}.

	\begin{wrapfigure}{R}{0.58\textwidth}
		\vspace{-20px}
		\begin{minipage}[h]{0.58\textwidth}
			\begin{table}[H]
				\scriptsize
				\setlength\tabcolsep{2pt}
				\centering
				\caption{Cross-domain results on META-DATASET (accuracy (\%)).}
				\begin{tabular}{c|cc|cc|cc|cc|cc}
					\toprule
					{Baseline}   & \multicolumn{2}{c}{{MatchNet}} & \multicolumn{2}{c}{{ProtoNet}} & \multicolumn{2}{c}{fo-MAML} & \multicolumn{2}{c}{fo-Proto-MAML} & \multicolumn{2}{c}{P>M>F} \\ \midrule
					\TheName{}   & w/o  &           w/            & w/o  &           w/            & w/o  &          w/          & w/o  &             w/             & w/o  &         w/         \\ \midrule
					ILSVRC     & 45.0 &      \textbf{51.1}      & 50.5 &      \textbf{52.3}      & 45.5 &    \textbf{54.1}     & 49.5 &       \textbf{54.3}        & 77.0 &   \textbf{78.6}    \\ \midrule
					Omniglot    & 52.2 &     \textbf{54.6 }      & 59.9 &      \textbf{61.2}      & 55.5 &    \textbf{63.7}     & 63.3 &       \textbf{69.8}        & 91.7 &   \textbf{93.3}    \\ \midrule
					Aircraft    & 48.9 &      \textbf{51.5}      & 53.1 &      \textbf{54.9}      & 56.2 &    \textbf{64.9}     & 55.9 &       \textbf{61.5 }       & 89.7 &   \textbf{91.1}    \\ \midrule
					Birds     & 62.2 &      \textbf{66.8}      & 68.7 &      \textbf{68.9}      & 63.6 &    \textbf{69.9}     & 68.6 &            68.6            & 92.9 &   \textbf{94.0}    \\ \midrule
					Textures    & 64.1 &      \textbf{67.6}      & 66.5 &     \textbf{68.4 }      & 68.0 &    \textbf{ 72.3}    & 66.4 &       \textbf{69.4}        & 86.9 &   \textbf{87.5}    \\ \midrule
					Quick Draw   & 42.8 &     \textbf{46.7 }      & 48.9 &      \textbf{50.0}      & 43.9 &    \textbf{48.5}     & 51.5 &       \textbf{53.1}        & 80.2 &   \textbf{83.3}    \\ \midrule
					Fungi     & 33.9 &      \textbf{36.4}      & 39.7 &     \textbf{ 40.9 }     & 32.1 &    \textbf{40.6}     & 39.9 &       \textbf{43.7}        & 78.2 &   \textbf{80.1}    \\ \midrule
					VGG Flower   & 80.1 &     \textbf{84.9 }      & 85.2 &      \textbf{88.0}      & 81.7 &    \textbf{90.4}     & 87.1 &       \textbf{91.0}        & 95.7 &   \textbf{96.8}    \\ \midrule
					Traffic Signs & 47.8 &      \textbf{49.5}      & 47.1 &      \textbf{48.6}      & 50.9 &    \textbf{52.2 }    & 48.8 &       \textbf{51.5}        & 89.8 &   \textbf{94.0}    \\ \midrule
					MS COCO    & 34.9 &      \textbf{40.1}      & 41.0 &      \textbf{42.4}      & 35.3 &    \textbf{43.5}     & 43.7 &       \textbf{48.9 }       & 64.9 &   \textbf{68.4}    \\ \hline
				\end{tabular}
				\label{tab:metadataset}
			\end{table}
		\end{minipage}
	\end{wrapfigure}

	\subsection{Cross-Domain Few-Shot Image Classification Performance}\label{app:metadataset}
	To show that \TheName{} is problem-agnostic, 
	we provide learner-agnostic improvement on large-scale cross-domain few-shot image classification problem, obtained on
	META-DATASET \citep{triantafillou2020meta}. 
	Table~\ref{tab:metadataset} shows the results. 
	The backbone and setting of P>M>F \citep{hu2022pushing} is different with the other baselines \citep{triantafillou2020meta}, so they are not comparable across baselines. 
	\TheName{} is introduced with the same setting as Section~\ref{sec:exp-fscv} (inner-task sampling $K=1$ and $\pi_\kappa(\cD^{\text{tr}}_\tau\cup\cD^{\text{val}}_\tau)=\cD^{\text{tr}}_\tau$, $\phi(a,b)=1-\nicefrac{a\cdot b}{\Vert a\Vert\Vert b\Vert}$ (cosine distance) and $\lambda=0.1$.). 
	Note that for P>M>F, 
	\TheName{} is integrated into the meta-training phase, and all other phases remain unchanged. 
	As shown, ConML brings consistent improvement.
	
	\subsection{Model Analysis}
	
	We show
	\TheName{} does not require much efforts on tuning hyperparameters.
	Furthermore, better performance can be obtained through hyperparameter optimization for specific meta-learners. 
	In this Section we show the impact of key \TheName{} settings: (1) the number of subset samples $K$, 
	which influences the model's complexity, and (2) the contrastive loss, including the distance function $\phi$, 
	the weighting factor $\lambda$, and the use of InfoNCE as a replacement for $(d^{\text{in}} - d^{\text{out}})$. 
	
	\subsubsection{Effect of the Number of Subset Samples $K$}
	\label{sec:k}
	
	Table~\ref{tab:k} presents the results of varying the number of subset samples $K$. 
	Starting from $K=1$, we observe moderate performance growth as $K$ increases, while memory usage grows linearly with $K$. 
	Notably, there is a significant discrepancy in both performance and memory (approximately $\sim2\times$) between the configurations without \TheName{} and with $K=1$. 
	However, $K$ has a negligible impact on time efficiency, assuming sufficient memory, as the processes are independent and can be executed in parallel.
	
	\subsubsection{The Design of Contrastive Loss}
	\label{sec:discus-sample}
	
	Here, we explore various design factors of the contrastive loss.
	\TheName{} optimizes the following objective: 
	$\mL_{\text{\TheName}} = \mL_e + \lambda \mL_c$.
	In the previous sections, to highlight our motivation and perform a decoupled analysis, we used the naive contrastive loss 
	$\mL_c = d^{\text{in}} - d^{\text{out}}$, 
	with the natural cosine distance  $\phi(x,y)$. 
	Here, we consider distance function $\phi$ as Euclidean distance, 
	contrastive loss $\mL_c$ in the form of InfoNCE \cite{oord2018representation}, 
	varying contrastive weight $\lambda$ in a wide range. More details are 
	in Appendix~\ref{app:hyper}. 
	
	\begin{minipage}[h]{0.49\textwidth}
		\vspace{-15px}
		\begin{table}[H]
			\centering
			\scriptsize
			\setlength\tabcolsep{2pt}
			\caption{The effect of subset sampling number $K$.}
			\begin{tabular}{c|c|c|c|c|c|c}
				\toprule
				& & w/o& K=1  & 4  & 16 & 32 \\
				\midrule
				\multirow{3}{*}{\makecell{MAML\\w/ \TheName{}}} & Acc.(\%) & 48.75& 56.25  & 56.08& \textbf{57.59}   & 57.33 \\\cmidrule{2-7}
				& Mem.(MB) & 1331  & 2801  & 3011  & 4103 & 5531 \\\cmidrule{2-7}
				&Time (relative)& 1$\times$  & 1.1$\times$ & 1.1$\times$  & 1.1$\times$  & 1.1$\times$    \\\midrule
				\multirow{3}{*}{\makecell{ProtoNet\\w/ \TheName{}}} & Acc.(\%) & 48.90 & 51.03  & 52.04  &52.34 &\textbf{52.48 } \\\cmidrule{2-7}
				& Mem.(MB) & 7955 &14167 & 15175&19943 &26449 \\\cmidrule{2-7}
				&Time (relative)& 1$\times$   & 1.2$\times$  & 1.2$\times$  & 1.2$\times$ & 1.2$\times$  \\
				\bottomrule
			\end{tabular}
			\label{tab:k}
		\end{table}
	\end{minipage}
	\hspace{0.02\textwidth}
	\begin{minipage}[h]{0.49\textwidth}
		\begin{figure}[H]
			\centering
			\subfigure[MAML w/ \TheName{}.]{
				\includegraphics[width=0.47\textwidth]{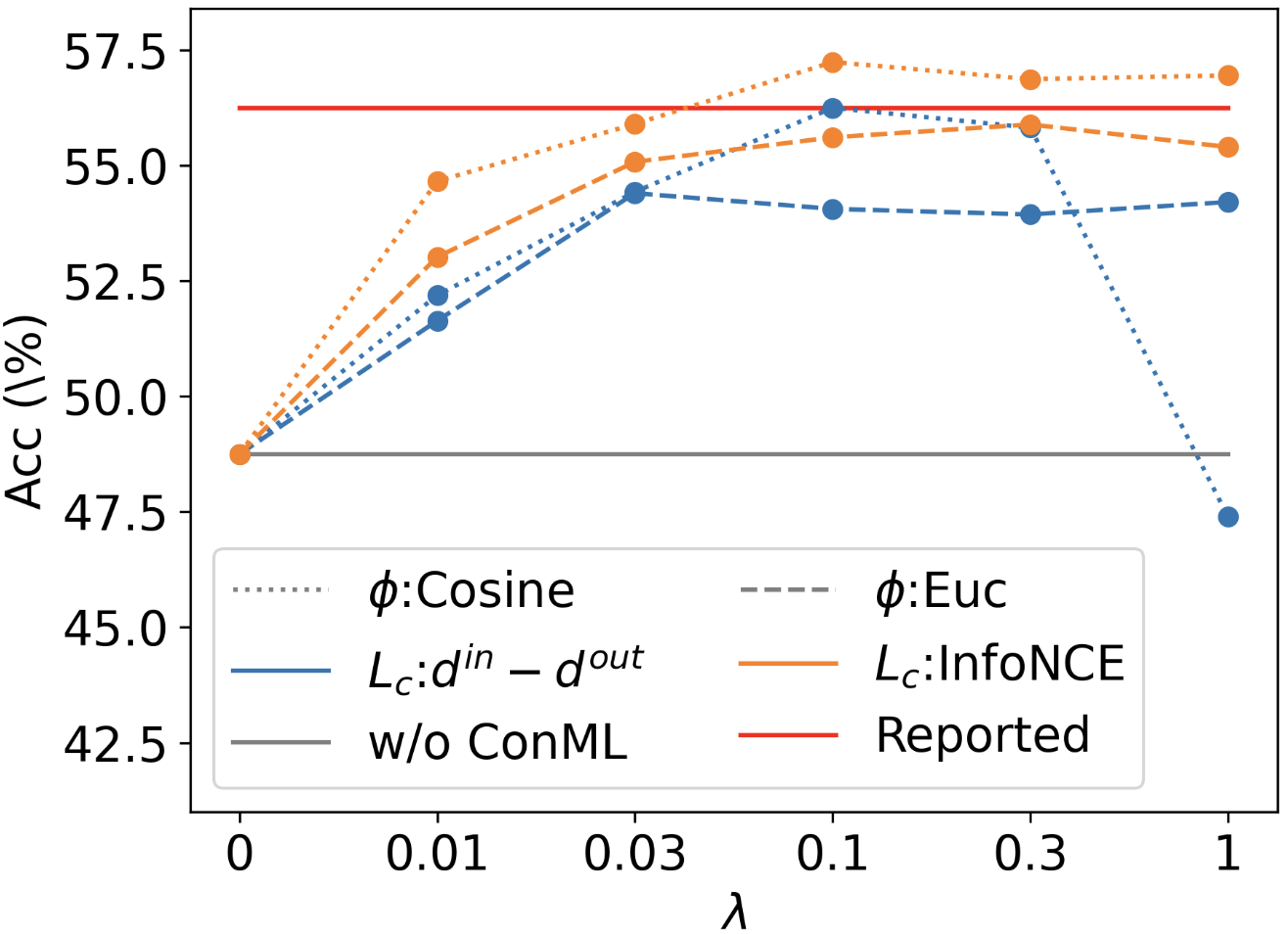}}
			\subfigure[ProtoNet w/ \TheName{}.]{
				\includegraphics[width=0.47\textwidth]{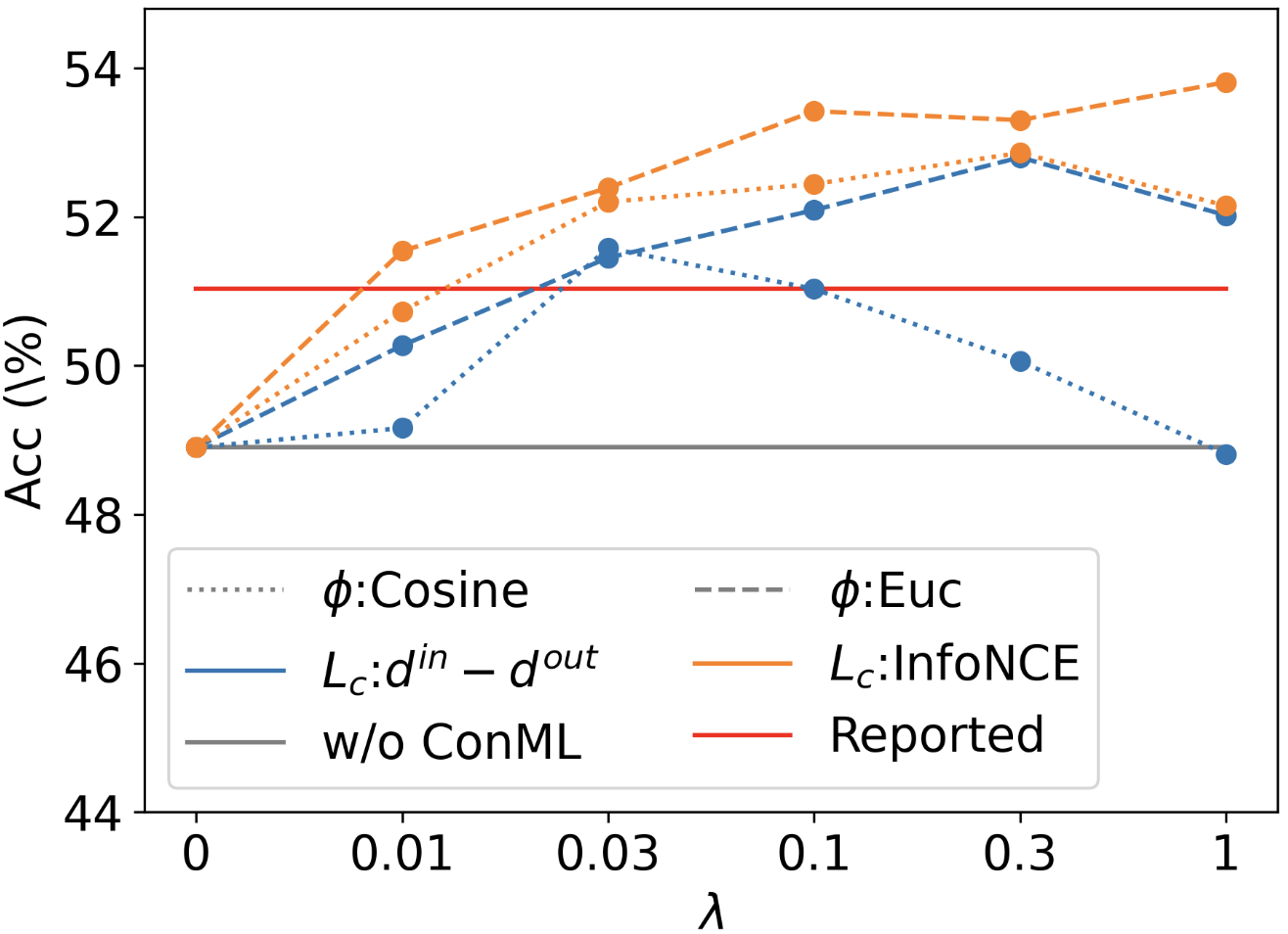}}
			\caption{The effect of distance function $\phi$, contrastive loss form $\mL_c$, contrastive weight $\lambda$ .}
			\label{fig:phi}
		\end{figure}
	\end{minipage}

	Figure~\ref{fig:phi} presents the results. We observe that \TheName{} can significantly improve the performance of meta-learners across a considerable range of $\lambda$, though setting $\lambda$ too high can lead to model collapse by overshadowing the original meta-learning objective. The choice of distance function varies between algorithms, with some performing better with specific functions. Additionally, InfoNCE outperforms the naive contrastive strategy, offering greater potential and reduced sensitivity to hyperparameters. 
	
	These findings suggest that we may not have yet reached the full potential of \TheName{}, and there are several promising directions for further improvement. 
	For instance, refining batch sampling strategies to account for task-level similarities or developing more advanced subset-sampling methods could enhance performance further \cite{liu2020adaptive,wang2024towards,wang2025robust}.
	We also notice that the matching between the chosen distance metric and model representation is the key to success. We can find that Euclidean distance performs much better than cosine in ProtoNet, since ProtoNet makes classification with Euclidean distance, and ConML contrasts the classifier's weights describing the model's behavior more precisely. Although cosine works generally, it would be interesting to tailor distance metrics for various parameter types (e.g., classifiers, MLPs, CNNs, GNNs, Transformers). It is also worth noting that a more delicate choice of distance metric is implied by Theorem~\ref{theory1}. See details in Appendix~\ref{app:theory}.
	
	In Appendix~\ref{sec:exp:ana}, we provide empirical results under synthetic dataset to understand (i) learning to learn with \TheName{} brings generalizable alignment and discrimination abilities; and (ii)
	alignment enhances fast-adaptation and discrimination enhances task-level generalizability.

	\subsection{ICL Performance}\label{sec:exp-icl}

	Following \citep{garg2022can}, we investigate \TheName{} on ICL by learning to learn synthetic functions including linear regression (LR), sparse linear regression (SLR), decision tree (DT) and 2-layer neural network with ReLU activation (NN). We train the GPT-2 \citep{radford2019language}-like transformer for each function with ICL and ICL w/ \TheName{} 
	respectively and compare the inference (meta-testing) performance.
	We follow the same model structure, data generation and training settings \citep{garg2022can}.
	More implementation details are provided in Appendix~\ref{app:exp-icl}.
	\begin{figure}[H]
		\vspace{-5pt}
		\centering
		\subfigure[LR.\label{fig:icl-linear}]{
			\includegraphics[width=0.235\textwidth]{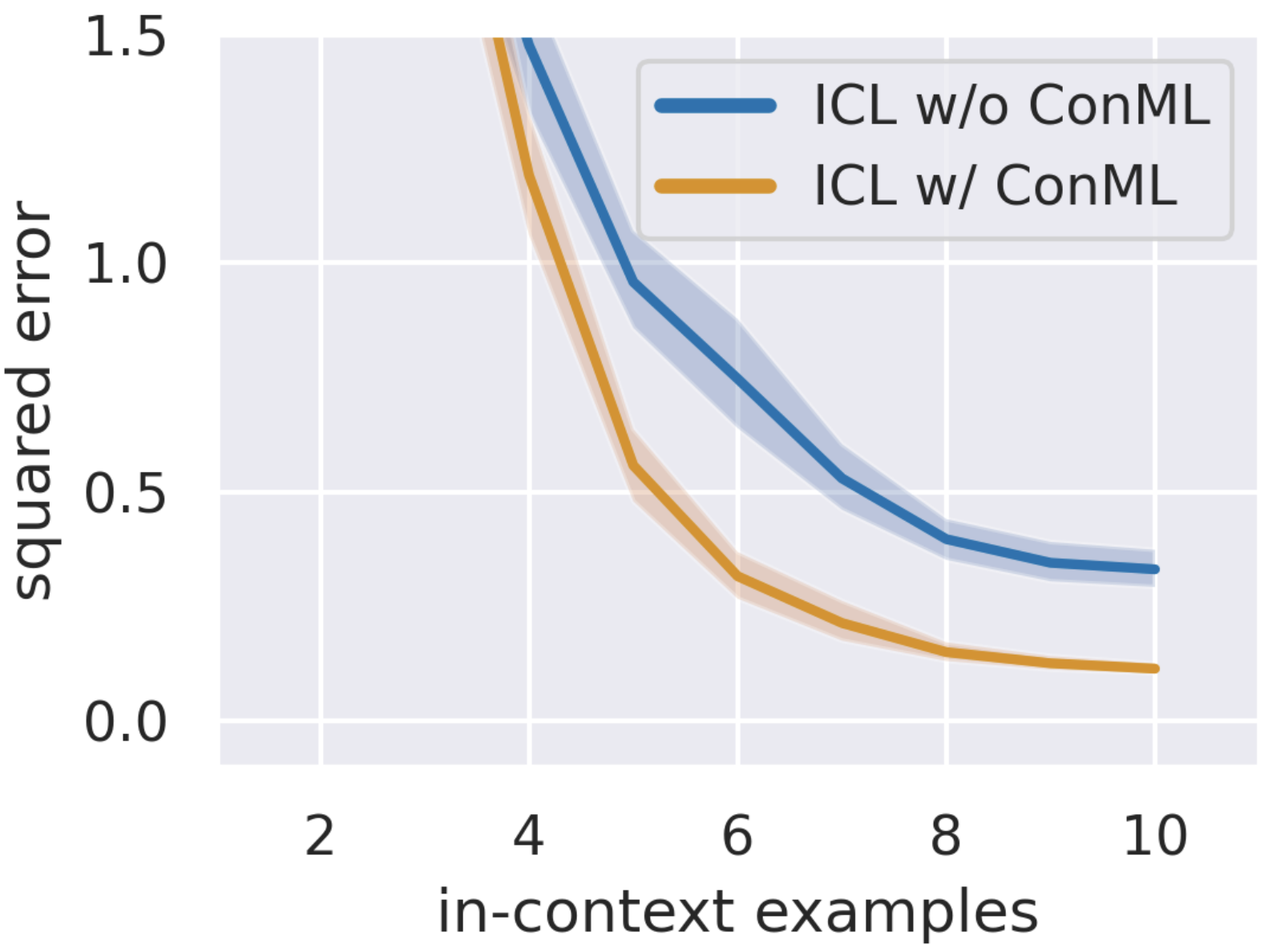}}
		\subfigure[SLR.\label{fig:icl-slr}]{
			\includegraphics[width=0.235\textwidth]{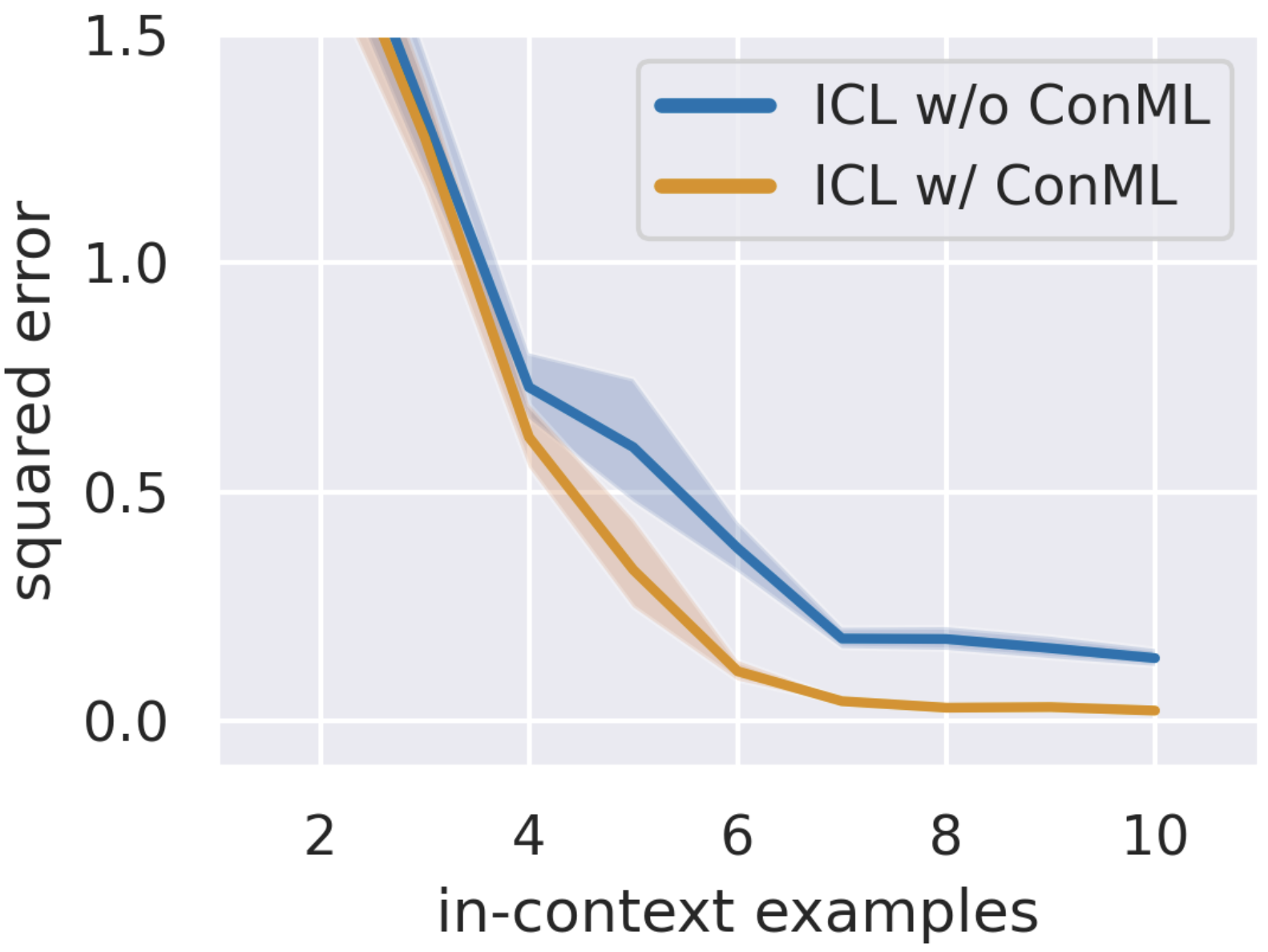}}
		\subfigure[DT.\label{fig:icl-dt}]{
			\includegraphics[width=0.235\textwidth]{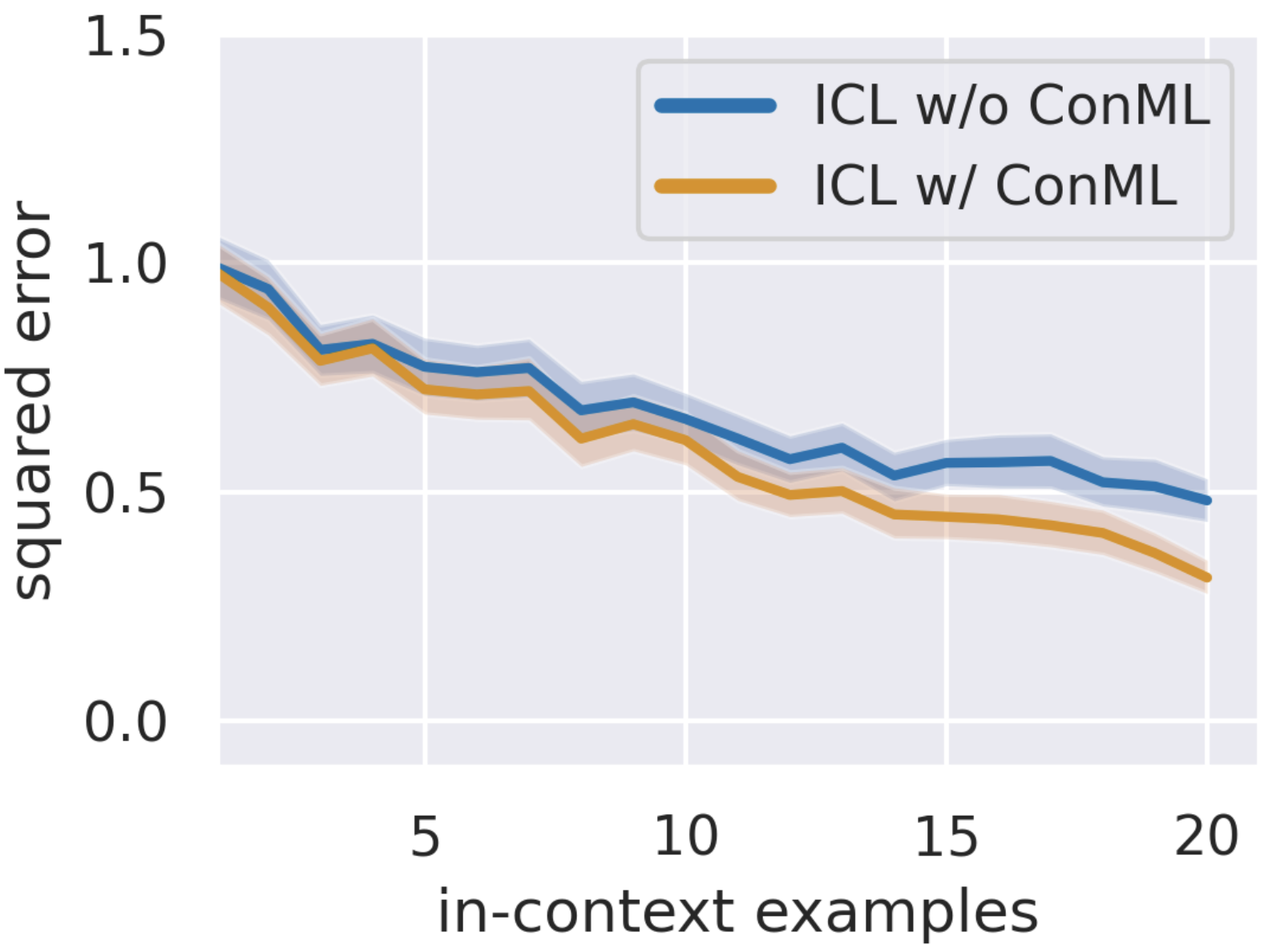}}
		\subfigure[NN.\label{fig:icl-nn}]{
			\includegraphics[width=0.235\textwidth]{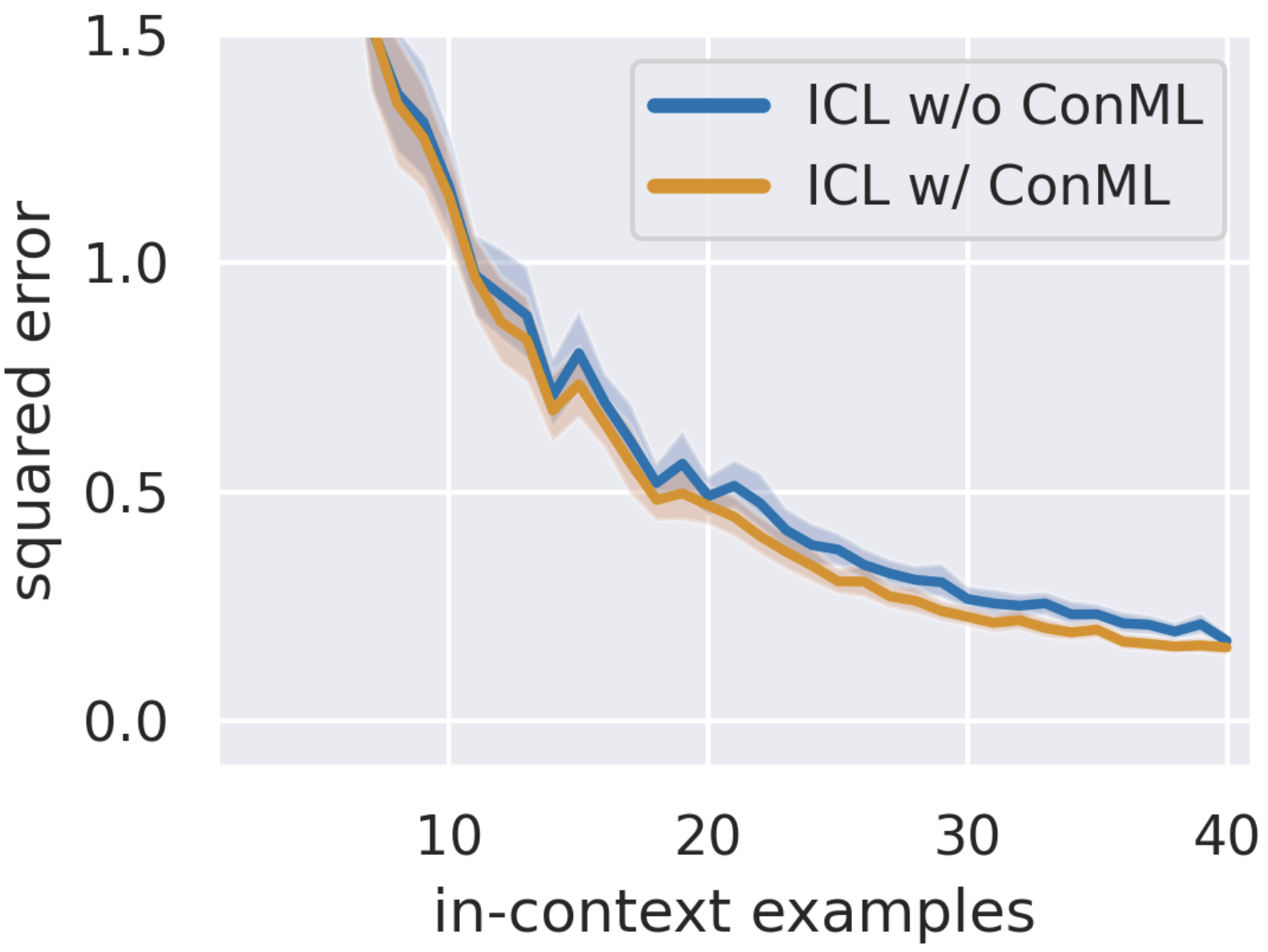}}
		\caption{Varying the number of in-context examples during inference of ICL.}
		\label{fig:icl}
	\end{figure}
	\begin{table}[h]
		\centering
		\vspace{-10pt}
		\scriptsize
		\caption{Relative minimal error (Rel. Min. Error) and spared example number to reach the same error (Shot Spare) comparing ICL w/ and w/o \TheName{}.}
		\begin{tabular}{c |c | c|c|c}
			\toprule
			\makecell{Function\\(max prompt len.)}&	 LR (10 shot)  & SLR (10 shot)&	DT (20 shot) & NN (40 shot)\\ \midrule
			Rel. Min. Error& $0.42\pm 0.09$&$0.49\pm.06$& $0.81\pm0.12$&$0.74\pm0.19$\\ \midrule
			Shot Spare& $-4.68\pm0.45$&$-3.94\pm0.62$&$-4.22\pm1.29$&$-11.25\pm2.07$\\ \bottomrule
		\end{tabular}
		\label{tab:icl}
		\vspace{-10px}
	\end{table}
	
	Figure \ref{fig:icl} shows the performance, where ICL w/ \TheName{} always makes more accurate predictions than ICL w/o \TheName{}.
	Table~\ref{tab:icl} shows the two values to show the effect \TheName{} brings to ICL: \textit{Rel. Min. Error} is ICL w/ \TheName{}'s minimal inference error given different number of examples, divided by ICL's;  and 
	\textit{Shot Spare} is when ICL w/ \TheName{} obtain an error no larger than ICL's minimal error, the difference between the corresponding example numbers.
	One can observe
	significant improvement.
	The effect of \TheName{} on ICL is without loss of generalizability to real-world applications like pretrained LLMs.

	\section{Relation with Existing Works}\label{sec:relate}
	
	\paragraph{Task-Identity in Meta-Learning}
	There are existing works attempt to leverage task-identity information into meta-learning, but no "counterpart" for \TheName{}.
	We discuss them in two categories: (i)
	The first category is using problem-specific information, while \TheName{} uses problem-agnostic information thus can be plugged-in these methods and brings improvement. They primarily focusing on few-shot image classification problem \citep{doersch2020crosstransformers,hiller2022rethinking,perera2024discriminative}, and require a static pool of base classes for meta-training and class-level alignment \citep{wang2016learning,fei2021melr,ye2022few,tian2022improving,wang2023robust}.
	These problem-specific approaches are limited by their focus on few-shot classification and cannot effectively handle dynamic or diverse tasks, nor can they generalize to other meta-learning problems beyond classification. As such, they are not directly comparable with \TheName{}. However, \TheName{} can be integrated into these methods. 
	Though they introduce new objectives other than validation loss by additional modules or steps, but they all work under/with the general mini-batch episodic training, either by replacing the steps to obtain $\mL_e$ in Algorithm~\ref{alg:episodic-training} with their steps to obtain episodic loss, or introduce additional steps outside Algorithm~\ref{alg:episodic-training}. 
	We demonstrate in Section~\ref{sec:exp-fscv} that incorporating \TheName{} leads to performance gains.
	(ii) The second category includes works that are learner-specific but not problem-specific. For example, \cite{gondal2021function} and \cite{mathieu2021contrastive} explore contrastive representations for neural processes. However, their methods are tightly coupled with specific meta-learners that involve explicit model representation vectors, which can be seen as special cases of \TheName{} within amortization-based meta-learners. 
	
	\paragraph{Contrastive Learning with Meta-Learning}
	Some studies involve both meta-learning and contrastive learning as key components, but they are not directly related to \TheName{}. 
	\cite{ni2021close} reformulates contrastive learning through meta-learning for better unsupervised learning, while \cite{zucchet2022contrastive} proposes an optimization-based meta-learner inspired by contrastive Hebbian learning in biology, which is not related to the contrastive learning used in unsupervised learning. \cite{lee2023self} introduces contrastive set representations for unsupervised meta-learning but does not integrate them with the general meta-learning framework or model. 

	\section{Conclusion, Limitations and Discussion}
	
	In this work, we propose \TheName{}, a universal, learner-agnostic contrastive meta-learning framework that emulates the alignment and discrimination capabilities integral to human fast learning, achieved through task-level contrastive learning in the model space. \TheName{} can be seamlessly integrated with meta-training procedure of existing meta-learners, by modifying the conventional mini-batch episodic training, and we provide specific implementations across a wide range of meta-learning algorithms. 
	Empirical results show that \TheName{} consistently and significantly enhances meta-learning performance by improving the meta-learner's fast-adaptation and task-level generalization abilities. Additionally, we explore in-context learning by reformulating it within the meta-learning paradigm, demonstrating how \TheName{} can be effectively integrated to boost performance. 
	
	The primary contribution of \TheName{} is offering a universal framework built and on the general meta-learning setting and training procedure, to reflects the inherency of alignment and discrimination as meta-training objective and the efficacy of learning to learn with contrasting model representation. 
	The cost of \TheName{} is additional training cost, as dicussed in Section~\ref{sec:k}, which is moderate but indelible under such framework.
	The current implementation of \TheName{} is relatively primitive, as discussed in Section~\ref{sec:discus-sample}, there are many directions for further improvement, such as optimizing sampling strategies, task-scheduling, refining the contrastive strategy and tailoring model representations and distance metrics.
	
	\section*{Acknowledgment}
	Y. Wang is sponsored by Beijing Nova Program. 
	Q. Yao is supported by National Natural Science Foundation of China 
	(under Grant No. 92270106) and Beijing Natural Science Foundation (under Grant No. 4242039). Y. Bian is supported by the National University of Singapore SoC (grant no: A-0010308-00-00).
	
	\bibliography{conml}
	\bibliographystyle{plain}

	\newpage
	\section*{NeurIPS Paper Checklist}
	
	\begin{enumerate}
		
		\item {\bf Claims}
		\item[] Question: Do the main claims made in the abstract and introduction accurately reflect the paper's contributions and scope?
		\item[] Answer: \answerYes{}
		\item[] Justification: In this work, we propose \TheName{}, a universal, learner-agnostic contrastive meta-learning framework that emulates the alignment and discrimination capabilities integral to human fast learning, achieved through task-level contrastive learning in the model space. \TheName{} can be seamlessly integrated with meta-training procedure of existing meta-learners, by modifying the conventional mini-batch episodic training, and we provide specific implementations across a wide range of meta-learning algorithms. 
		Empirical results show that \TheName{} consistently and significantly enhances meta-learning performance by improving the meta-learner's fast-adaptation and task-level generalization abilities. Additionally, we explore in-context learning by reformulating it within the meta-learning paradigm, demonstrating how \TheName{} can be effectively integrated to boost performance. 
		
		\item {\bf Limitations}
		\item[] Question: Does the paper discuss the limitations of the work performed by the authors?
		\item[] Answer: \answerYes{}
		\item[] Justification: The paper discusses the limitations of the work performed by the authors, as summarized in the last paragraph of the main text.
		
		\item {\bf Theory assumptions and proofs}
		\item[] Question: For each theoretical result, does the paper provide the full set of assumptions and a complete (and correct) proof?
		\item[] Answer:  \answerYes{}.
		\item[] Justification: For each theoretical result, the paper provides the full set of assumptions and a complete (and correct) proof, involving assumption and results in Section~\ref{sec:theory} and proof in Appendix \ref{app:theory}.
		
		\item {\bf Experimental result reproducibility}
		\item[] Question: Does the paper fully disclose all the information needed to reproduce the main experimental results of the paper to the extent that it affects the main claims and/or conclusions of the paper (regardless of whether the code and data are provided or not)?
		\item[] Answer: \answerYes{} 
		\item[] Justification:  The experiment setting and methods are fully described.

		\item {\bf Open access to data and code}
		\item[] Question: Does the paper provide open access to the data and code, with sufficient instructions to faithfully reproduce the main experimental results, as described in supplemental material?
		\item[] Answer:  \answerYes{} 
		\item[] Justification: Code is provided at \url{https://github.com/ovo67/ConML_Code}.

		\item {\bf Experimental setting/details}
		\item[] Question: Does the paper specify all the training and test details (e.g., data splits, hyperparameters, how they were chosen, type of optimizer, etc.) necessary to understand the results?
		\item[] Answer:  \answerYes{} 
		\item[] Justification: Most key details are directly mentioned and discussed, while the others can be figured out in the provided code.
		
		\item {\bf Experiment statistical significance}
		\item[] Question: Does the paper report error bars suitably and correctly defined or other appropriate information about the statistical significance of the experiments?
		\item[] Answer:  \answerYes{} 
		\item[] Justification: Most results are provided with standard derivation. Exclusions are Table~\ref{tab:metadataset} that requires too much source we could not afford to do multiple times, and neglecting for paper space limitation at Table~\ref{tab:k}.
		
		\item {\bf Experiments compute resources}
		\item[] Question: For each experiment, does the paper provide sufficient information on the computer resources (type of compute workers, memory, time of execution) needed to reproduce the experiments?
		\item[] Answer:  \answerNo{}, 
		\item[] Justification: Table~\ref{tab:k} has shown the detailed relative consumption comparing the proposed method and standard method.
		
		\item {\bf Code of ethics}
		\item[] Question: Does the research conducted in the paper conform, in every respect, with the NeurIPS Code of Ethics \url{https://neurips.cc/public/EthicsGuidelines}?
		\item[] Answer: \answerYes{} 
		\item[] Justification: The research conducted in the paper conforms, in every respect, with the NeurIPS Code of Ethics

		\item {\bf Broader impacts}
		\item[] Question: Does the paper discuss both potential positive societal impacts and negative societal impacts of the work performed?
		\item[] Answer:  \answerNA{}, 
		\item[] Justification: There is no societal impact of the work performed.
		
		\item {\bf Safeguards}
		\item[] Question: Does the paper describe safeguards that have been put in place for responsible release of data or models that have a high risk for misuse (e.g., pretrained language models, image generators, or scraped datasets)?
		\item[] Answer: \answerNA{}. 
		\item[] Justification: The paper poses no such risks.
		
		\item {\bf Licenses for existing assets}
		\item[] Question: Are the creators or original owners of assets (e.g., code, data, models), used in the paper, properly credited and are the license and terms of use explicitly mentioned and properly respected?
		\item[] Answer:  \answerYes{} 
		\item[] Justification: The creators or original owners of assets (e.g., code, data, models), used in the paper, are properly credited and are the license and terms of use explicitly mentioned and properly respected.
		
		\item {\bf New assets}
		\item[] Question: Are new assets introduced in the paper well documented and is the documentation provided alongside the assets?
		\item[] Answer:  \answerYes{} 
		\item[] Justification: Yes.
		
		\item {\bf Crowdsourcing and research with human subjects}
		\item[] Question: For crowdsourcing experiments and research with human subjects, does the paper include the full text of instructions given to participants and screenshots, if applicable, as well as details about compensation (if any)? 
		\item[] Answer: \answerNA{} 
		\item[] Justification: The paper does not involve crowdsourcing nor research with human subjects.
		
		\item {\bf Institutional review board (IRB) approvals or equivalent for research with human subjects}
		\item[] Question: Does the paper describe potential risks incurred by study participants, whether such risks were disclosed to the subjects, and whether Institutional Review Board (IRB) approvals (or an equivalent approval/review based on the requirements of your country or institution) were obtained?
		\item[] Answer:\answerNA{} 
		\item[] Justification: The paper does not involve crowdsourcing nor research with human subjects.
		
		\item {\bf Declaration of LLM usage}
		\item[] Question: Does the paper describe the usage of LLMs if it is an important, original, or non-standard component of the core methods in this research? Note that if the LLM is used only for writing, editing, or formatting purposes and does not impact the core methodology, scientific rigorousness, or originality of the research, declaration is not required.
		\item[] Answer: \answerNA{} 
		\item[] Justification: The core method development in this research does not involve LLMs as any important, original, or non-standard components.
		
	\end{enumerate}
	
	\clearpage
	
	\onecolumn
	
	\appendix

	\section{Complexity Analysis}\label{app:complex}
	We compare the relative complexity of computing the original meta-objective and the additional contrastive objective introduced by ConML. 
	\subsection{ICL}
	For ICL model like LLM, ConML does not obtain model representation by explicit model parameters, but by simply adding an additional token to the forward-pass ($u$ in \eqref{eq:repre-icl}). Which means pretraining a LLM with ConML only requires $K/n$ ($K$: subset sampling number, $n$: average sentence length, typically $K/n<<1$) times computation comparing with pretraining a LLM without ConML, regardless of the model size.
	\subsection{Typical Meta-Learners}
	For typical meta-learners,
	denote the model representation has $d$ parameters, i.e., $e\in R^d$. We discuss about the complexity of the original computation path $h=g(D;\theta) \rightarrow \mL_e$ and the additional computation path $h=g(D;\theta)\rightarrow \psi(h)\rightarrow \mL_c$ introduced by ConML. We consider giving a single input sample in 1-d vector, the complexity $O_{h\rightarrow \mL_e}$ to calculate the loss $\mL_e=\ell(h(x),y)$, and the complexity $O_{h\rightarrow  \psi(h)\rightarrow \mL_c}$ to calculate $\mL_c=d(\psi(h),\psi(h))$.
	\begin{itemize}
		\item For optimization-based, e.g., MAML, we have $d= |\theta|=|h|$. We consider $h$ as a $l$-layer MLP, with each average layer size $(|h|/l)^{1/2} * (|h|/l)^{1/2}$. With a single input sample, $O_{h\rightarrow \mL_e}=O(l*(|h|/l)^{3/2})$, $O_{h\rightarrow  \psi(h)\rightarrow \mL_c}=O(|d|)+O(|d|)=O(|h|)$.
		While $l <<|\theta|$, we have $O_{h\rightarrow \mL_e}>O_{h\rightarrow  \psi(h)\rightarrow \mL_c}$.
		
		\item For metric-based, e.g., ProtoNet, $\theta$ corresponds to the parameter in feature extractor like CNN. $h$ is the final classifier which makes prediction by Euclidean distance, which can be viewed as a linear classifier with parameter size in $N*|h|/N$. $d$ equals to the N (ways per task) times the dimension of the embedding of a each sample $|h|/N$, $d=|h|$. A sample $x\in R^{|h|/N}$.
		We have $O_{h\rightarrow \mL_e}=O(|h|^2/N)$, $O_{h\rightarrow  \psi(h)\rightarrow \mL_c}=O(d)+O(d)=O(|h|)$. With $|h|>>N$, $O_{h\rightarrow \mL_e}>O_{h\rightarrow  \psi(h)\rightarrow \mL_c}$.
		
		\item For amortization-based,e.g., Simple CNAPs. Denote $q$ as the dimension of task-adaptive parameters generated by hypernetwork $H_\theta(D)$. $|d|=q$, $O_{h\rightarrow\psi(h)\rightarrow \mL_c}=O(d)+O(d)=O(q)$. Consider l layers in main-network modulated by $H_\theta(D)$ feature-wisely, the $q=\sqrt{|h|/l}$, $O_{h\rightarrow \mL_e}=O(l*(|h|/l)^{3/2})+O(q/l)$. With $l<<|h|$, we have $O_{h\rightarrow \mL_e}>O_{h\rightarrow  \psi(h)\rightarrow \mL_c}$.
	\end{itemize}
	
	To summarize, for ICL model like LLMs, the complexity to pretrain with ConML is $\frac{n+K}{n}\approx1$ times the complexity to pretrain without ConML. For typical meta-learners, the additional introduced objective in ConML is comparably less than the complexity of the original meta-training objective, which verifies the empirical computation cost presented in Table~\ref{tab:k} in main text.

	\clearpage

	\section{Provable Benefits for Generalization}
	\label{app:theory}
	Here we first provide the proof of Lemma~\ref{lem:1} which shows $U_{p(\tau)}(\theta)$ is an upper bound of the excess risk of meta-learning $\Delta\epsilon_{p(\tau)}(\theta)$, and then the proof of Theorem~\ref{theory1} which shows minimizing contrastive meta-objective is minimizing $U_{p(\tau)}(\theta)$.

	We need two preliminary results:
	\begin{lemma}
		[Upper Bound from \cite{maurer2016benefit}]
		$\forall \theta$, $\Delta\epsilon_{p(\tau)}(\theta) \leq U_{p(\tau)}(\theta)$, 
		where
		\begin{align*}
			\Phi_{p(\tau)}(\theta)=C_1\sqrt{{E_{\tau\sim p(\tau)}}{E_{(x,y)\sim \tau}}[||\psi(g((x,y);\theta))||^2]}+C_2,
		\end{align*}
		with $C_1,C_2>0, \frac{dC_1}{dg}=\frac{dC_2}{dg}=0$ .
	\end{lemma}
	
	\begin{lemma}
		[Universal Approximation of MLP from \cite{hornik1991approximation}]\label{tho:approx}
		Let $\sigma: \mathbb{R} \to \mathbb{R}$ be a non-constant, bounded, and continuous function.
		Let $K$ be a compact subset of $\mathbb{R}^n$. The set of real-valued continuous functions on $K$ is denoted by $C(K)$.
		For any function $f \in C(K)$ and for any error tolerance $\delta > 0$, there exists an integer $N$ (the number of neurons in the hidden layer), and real constants $v_i, b_i \in \mathbb{R}$ and vectors $w_i \in \mathbb{R}^n$ for $i=1, \dots, N$, such that we can define the MLP output function $F: K \to \mathbb{R}$ as:
		\[
		F(x) = \sum_{i=1}^{N} v_i \sigma(w_i^T x + b_i)
		\]
		which satisfies
		$\forall x \in K$
		and
		$|f(x) - F(x)| < \delta$.
	\end{lemma}
	
	\subsection{Proof of Lemma~\ref{lem:1}}
	
	\begin{proof}
		We use the Universal Approximation of MLP to bound the $\psi$ in $\Phi_{p(\tau)}(\theta)$.
		
		For the model space $\cH$ (recall $h = g(\cD; \theta) \in \cH$) is closed and bounded by the meta-learner's hypothesis, which is a compact set on $\mathbb{R}^{|\cH|}$,
		we can apply Theorem~\ref{tho:approx} on $\psi$ in $\Phi_{p(\tau)}(\theta)$.
		There exists two-layer MLP $F$ with a nonlinear activation function $\sigma$. Then
		\begin{align}\label{eq:app}
			\forall X_g \in \cH, \quad|\psi(X_g)-F(X_g)|<\delta,
		\end{align}
		where $X_g$ represents the parameter vector learned by $g(;\theta)$ from sample $(x,y)$.

		As $X_g \in \cH$, $F(X_g)\in \mathbb{R}$ and $||F(X_g)||=||\psi(X_g)||\leq||X_g||$ by definition,
		we have
		\begin{align}\nonumber
			\exists w_1,\cdots,w_N \in \mathbb{R}^{|\cH|},\sum_{k=1}^{N}||w_k||^2\leq N, F(X_g)=\sum_{k=1}^{N}\sigma(\langle w_k,X_g\rangle)
		\end{align}
		
		As $\delta$ can be arbitrarily small by selecting large enough $N$, we approximately rewrite \eqref{eq:app} as $\psi(X_g)=F(X_g)$ which is not completely rigorous but no harm to our proof. We have
		\begin{align}\nonumber
			{E_{\tau\sim p(\tau)}}{E_{(x,y)\sim \tau}}[||\psi(g((x,y);\theta))||^2]={E}_{X\sim P_{g(\theta)(\tau,x,y)}}[\sum_{k}\alpha(\langle w_k,X_g\rangle)^2],
		\end{align}
		where  $P_{g(\theta)(\tau,x,y)}$ is the distribution of $X_g$ output by meta-learner $g(;\theta)$ on defined $p(\tau),x,y$.
		
		Let the activation function $\sigma$ has Lipschitz constant $L_{\sigma}$ and $\sigma(0)=0$. We have 
		\begin{align}\nonumber
			{E}_{X\sim P_{g(\theta)(\tau,x,y)}}[\sum_{k}\alpha(\langle w_k,X_g\rangle)^2]\leq& L_{\sigma}^2 \sum_{k=1}^{N}||w_k||^2 {E}_{X\sim P_{g(\theta)(\tau,x,y)}} [\langle \frac{w_k}{||w_k||},X_g\rangle^2]\\
			\leq&  L_{\sigma}^2 N \sup_{||v||\leq 1} {E}_{X\sim P_{g(\theta)(\tau,x,y)}} [\langle v,X\rangle^2].\nonumber
		\end{align}
		So we have 
		\begin{align}\nonumber
			\Phi_{p(\tau)}(\theta)=C_1\sqrt{{E_{\tau\sim p(\tau)}}{E_{(x,y)\sim \tau}}[||\psi(g((x,y);\theta))||^2]}+C_2\\
			\leq C_1\sqrt{L_{\sigma}^2N\sup_{||v||\leq 1}{E_{\tau\sim p(\tau)}} {E_{(x,y)\sim \tau}}[\langle v,g(\{(x,y)\};\theta)\rangle^2]}+C_2=U_{p(\tau)}(\theta).\nonumber
		\end{align}
		So $U_{p(\tau)}(\theta)$ is a upper bound of he the excess risk of a meta-learner $\Delta\epsilon_{p(\tau)}(\theta)$.
	\end{proof}
	

	\subsection{Proof of Theorem~\ref{theory1}}
	We need
	assuming $\forall \cD, ||g(\cD;\theta)||=1$, and choosing the distance function $\phi$ in $\mL_c$ to be $\phi(a,b)=-\frac{a\cdot b}{\Vert a\Vert\Vert b\Vert}$ for $d^{in}$ and $\phi(a,b)=-(\frac{a\cdot b}{\Vert a\Vert\Vert b\Vert})^2$ for $d^{out}$.
	We will discuss about the rationality of such assumption and choice after proof.
	\begin{proof}
		On the one hand, we have
		\begin{align}\nonumber
			U_{p(\tau)}(\theta)&=C_1\sqrt{L_{\sigma}^2N\sup_{||v||\leq 1}{E_{\tau\sim p(\tau)}} {E_{(x,y)\sim \tau}}[\langle v,g(\{(x,y)\};\theta)\rangle^2]}+C_2\\\nonumber
			&=C_1\sqrt{L_{\sigma}^2 N \sup_{||v||\leq 1} {E}_{X\sim P_{g(\theta)(\tau,x,y)}} [\langle v,X\rangle^2]}+C_2,~s.t. ||X||=1\\\nonumber
			&\geq C_1\sqrt{\frac{L_{\sigma}^2 N}{|\cH|}} +C_2,
		\end{align}
		where the minimum of $U_{p(\tau)}(\theta)$
		\begin{align}\nonumber
			U^*_{p(\tau)}(\theta)=C_1\sqrt{\frac{L_{\sigma}^2 N}{|\cH|}} +C_2
		\end{align}
		is achieved if and only if
		\begin{align}\nonumber
			\sup_{||v||\leq 1, s.t. ||X||=1 } {E}_{X\sim P_{g(\theta)(\tau,x,y)}} [\langle v,X\rangle^2]= \frac{1}{|\cH|},
		\end{align}
		which is achieved if and only if 
		\begin{align}\nonumber
			\forall ||X||=1, P_{g(\theta)(\tau,x,y)}(X)=\frac{\Gamma(\frac{|\cH|}{2})}{2\pi^{\frac{|\cH|}{2}}},
		\end{align}
		i.e., $X$ uniformly distribute on the unit sphere in $\mathbb{R}^{|\cH|}$.
		
		On the other hand, by definition we have
		\begin{align}\nonumber
			\mL_c=&d^{in}-d^{out}\\\nonumber
			=&{E}_{X_{\tau,\kappa}\sim P_{g(\theta,\pi_\kappa)(\tau,x,y,\kappa)}}[-\frac{\langle X_{\tau,\kappa},X_{\tau,\kappa'}\rangle}{||X_{\tau,\kappa}||||X_{\tau,\kappa'}||}+(\frac{\langle X_{\tau},X_{\tau'}\rangle}{||X_{\tau}||||X_{\tau'}||})^2]\\\nonumber
			=&E_{\tau\sim P_{g(\theta)(\tau)}}~[E_{X_{\tau,\kappa}\sim P_{g(\theta,\pi_\kappa)(x,y,\kappa\mid\tau)}}[-\frac{\langle X_{\tau,\kappa},X_{\tau,\kappa'}\rangle}{||X_{\tau,\kappa}||||X_{\tau,\kappa'}||}]~+(\frac{\langle X_{\tau},X_{\tau'}\rangle}{||X_{\tau}||||X_{\tau'}||})^2]\\\nonumber
			=&-E_{\tau\sim P_{g(\theta)(\tau)},X_{\tau,\kappa}\sim P_{g(\theta,\pi_\kappa)(x,y,\kappa\mid\tau)}}[\frac{\langle X_{\tau,\kappa},X_{\tau,\kappa'}\rangle}{||X_{\tau,\kappa}||||X_{\tau,\kappa'}||}]+E_{\tau\sim P_{g(\theta)(\tau)}}[(\frac{\langle X_{\tau},X_{\tau'}\rangle}{||X_{\tau}||||X_{\tau'}||})^2].
		\end{align}
		For arbitrary subset sampling strategy $\pi_\kappa$, we have
		\begin{align}\nonumber
			\min_{\theta,s.t. ||X||=1x}\mL_c\geq-1+\frac{1}{|\cH|},
		\end{align}
		where the minimum of $\mL_c$
		\begin{align}\nonumber
			\mL^*_c=-1+\frac{1}{|\cH|},
		\end{align}
		is achieved if and only if 
		\begin{align}\nonumber
			E_{\tau\sim P_{g(\theta)(\tau)},X_{\tau,\kappa}\sim P_{g(\theta,\pi_\kappa)(x,y,\kappa\mid\tau)}}[\frac{\langle X_{\tau,\kappa},X_{\tau,\kappa'}\rangle}{||X_{\tau,\kappa}||||X_{\tau,\kappa'}||}]=1,\\\nonumber
			E_{\tau\sim P_{g(\theta)(\tau)}}[(\frac{\langle X_{\tau},X_{\tau'}\rangle}{||X_{\tau}||||X_{\tau'}||})^2]=\frac{1}{|\cH|},
		\end{align}
		which is achieved if and only if 
		\begin{align}\nonumber
			\forall ||\tau||=1,P_{g(\theta)}(\tau)=\frac{\Gamma(\frac{|\cH|}{2})}{2\pi^{\frac{|\cH|}{2}}},
			P_{g(\theta,\pi_\kappa)(x,y,\kappa\mid\tau)}(X\mid\tau)=\delta_\tau(X),
		\end{align}
		where $\delta_\tau$ is the Dirac-delta function centered on $\tau$.
		Then $\forall ||X||=1,$
		\begin{align}\nonumber
			P_{g(\theta,\pi_\kappa)(\tau,x,y,\kappa)}(X)=&\int_{||\tau||=1}P_{g(\theta)}(\tau) P_{g(\theta,\pi_\kappa)(x,y,\kappa\mid\tau)}(X\mid \tau)~d\tau\\\nonumber
			=&\int_{||\tau||=1}\frac{\Gamma(\frac{|\cH|}{2})}{2\pi^{\frac{|\cH|}{2}}}\delta_\tau(X)~d\tau\\
			=&\frac{\Gamma(\frac{|\cH|}{2})}{2\pi^{\frac{|\cH|}{2}}}\nonumber
		\end{align}
		
		Combining both hands, we have
		$$\theta^*_{\mL_c}=\arg\min_{\theta}\mL_c(g(;\theta),p(\tau))$$
		$$\Rightarrow \forall ||X||=1, P_{g(\theta,\pi_\kappa)(\tau,x,y,\kappa)}(X)=\frac{\Gamma(\frac{|\cH|}{2})}{2\pi^{\frac{|\cH|}{2}}}$$
		$$\Leftrightarrow U_{p(\tau)}(\theta^*_{\mL_c})=C_1\sqrt{\frac{L_{\sigma}^2 N}{|\cH|}} +C_2=\min_{\theta}U_{p(\tau)}(\theta)$$

	\end{proof}
	
	Now we discuss the implication of 
	assuming $\forall \cD, ||g(\cD;\theta)||=1$, and choosing the distance function $\phi$ in $\mL_c$ to be $\phi(a,b)=-\frac{a\cdot b}{\Vert a\Vert\Vert b\Vert}$ for $d^{in}$ and $\phi(a,b)=-(\frac{a\cdot b}{\Vert a\Vert\Vert b\Vert})^2$ for $d^{out}$.
	$||g(\cD;\theta)||=1$ can be viewed as a regularization on model weights of $h$ that prevents trivial solution and enhances generalization.
	
	The choice of $\phi$ is more inspiring: comparing with ordinary cosine distance for both $d^{in}$ and $d^{out}$, above form modifies $d^{out}$. For maximizing $d^{out}$, if we also choose ordinary cosine distance $\phi(a,b)=-\frac{a\cdot b}{\Vert a\Vert\Vert b\Vert}$, given a pair of tasks, it is optimized if and only if the two tasks are "opposite", i.e., still coupled, while the modified $-(\frac{a\cdot b}{\Vert a\Vert\Vert b\Vert})^2$ is maximized if and only if the two tasks are "orthogonal", i.e., decoupled. The latter form is preferred to be more reasonable. However, this understanding has only come to our mind after our work, so we have not implemented in our experiments. This could be studied as a future direction.

	\clearpage

	\section{Specifications of Meta-Learning with \TheName{}}\label{app:algs}
	Here, we provide the specific algorithm process of representative implementation \TheName{}, including 
	the universal framework of \TheName{} (Algorithm~\ref{alg:genera1}), the most efficient implementation of \TheName with $K=1$ and $\pi_\kappa(\cD^{\text{tr}}_\tau\cup\cD^{\text{val}}_\tau)=\cD^{\text{tr}}_\tau$ (Algorithm~\ref{alg:k1}), training ICL model with \TheName{} (Algorithm~\ref{alg:icl}), MAML w/ \TheName{} (Algorithm~\ref{alg:mamlc}), Reptile w/ \TheName{} (Algorithm~\ref{alg:reptilec}), SCNAPs w/ \TheName{} (Algorithm~\ref{alg:hyperc}), ProtoNet w/ \TheName{} (Algorithm~\ref{alg:protoc}).
	\begin{algorithm}[H]
		\caption{\TheName{}.}
		\label{alg:genera1}
		\begin{algorithmic}
			\STATE {\bfseries Input:} Task distribution $p(\tau)$, batch size $B$, inner-task sample times $K$ and sampling strategy $\pi_\kappa$.
			\WHILE {Not converged}
			\STATE Sample a batch of tasks $\bm{b}\sim p^B(\tau)$.
			\FOR { All $\tau\in \bm{b}$}
			\FOR {$k=1,2,\cdots,K$}
			\STATE Sample $\kappa_k$ from $\pi_\kappa(\cD^{\text{tr}}_\tau\cup\cD^{\text{val}}_\tau)$;
			\STATE Get model representation $\bm{e}^{\kappa_k}_{\tau}=  \psi(g(\kappa_k;\theta))$;
			\ENDFOR
			\STATE Get model representation $\bm{e}^*_{\tau}=  \psi(g(\cD^{\text{tr}}_\tau\cup\cD^{\text{val}}_\tau;\theta))$;
			\STATE Get inner-task distance $d^{\text{in}}_{\tau}$ by \eqref{eq:ind};
			\STATE Get task-specific model $h_{\tau}=g(\cD^{\text{tr}}_{\tau};\theta)$;
			\STATE Get validation loss $\mL(\cD^{\text{val}}_{\tau};h_\tau)$;
			\ENDFOR
			\STATE Get $d^{\text{in}}=\frac{1}{B}\sum_{\tau\in\bm{b}}d^{\text{in}}_\tau$ and $d^{\text{out}}$ by \eqref{eq:xd};
			\STATE Get loss $\mL_{\text{\TheName}}$ by \eqref{eq:obj};
			\STATE Update $\theta$ by $\theta\leftarrow\theta-\nabla_\theta \mL$.
			\ENDWHILE
		\end{algorithmic}
	\end{algorithm}

	\begin{algorithm}[ht]
		\caption{ ConML ($K=1$).}
		\label{alg:k1}
		\begin{algorithmic}
			\STATE {\bfseries Input:} Task distribution $p(\tau)$, batch size $B$ (inner-task sample times $K=1$ and sampling strategy $\pi_\kappa(\cD^{\text{tr}}_\tau\cup\cD^{\text{val}}_\tau)=\cD^{\text{tr}}_\tau$).
			\WHILE {Not converged}
			\STATE Sample a batch of tasks $\bm{b}\sim p^B(\tau)$.
			\FOR { All $\tau\in \bm{b}$}
			\STATE Get task-specific model $h_{\tau}=g(\cD^{\text{tr}}_{\tau};\theta)$, and model representation $\bm{e}^{\kappa_k}_{\tau}=  \psi(g(\kappa_k;\theta))$;
			\STATE Get model representation $\bm{e}^*_{\tau}=  \psi(g(\cD^{\text{tr}}_\tau\cup\cD^{\text{val}}_\tau;\theta))$;
			\STATE Get inner-task distance $d^{\text{in}}_{\tau}$ by \eqref{eq:ind};
			\STATE Get validation loss $\mL(\cD^{\text{val}}_{\tau};h_{\tau})$;
			\ENDFOR
			\STATE Get $d^{\text{in}}=\frac{1}{B}\sum_{\tau\in\bm{b}}d^{\text{in}}_\tau$ and $d^{\text{out}}$ by \eqref{eq:xd};
			\STATE Get loss $\mL_{\text{\TheName}}$ by \eqref{eq:obj};
			\STATE Update $\theta$ by $\theta\leftarrow\theta-\nabla_\theta \mL$.
			\ENDWHILE
		\end{algorithmic}
	\end{algorithm}

	\begin{algorithm}[ht]
		\caption{ICL with \TheName{} (ICL w/ \TheName{}).}
		\label{alg:icl}
		\begin{algorithmic}
			\STATE {\bfseries Input:} Task distribution $p(\tau)$, batch size $B$, inner-task sample times $K$ and sampling strategy $\pi_\kappa$, dummy input $u$ (probe).
			\WHILE {Not converged}
			\STATE Sample a batch of tasks $\bm{b}\sim p^B(\tau)$.
			\FOR { All $\tau\in \bm{b}$}
			\FOR {$k=1,2,\cdots,K$}
			\STATE Sample $\kappa_k$ from $\pi_\kappa(\cD_\tau)$;
			\STATE Get $\bm{e}^{\kappa_k}_{\tau}= g([\vec{\kappa_k},u];\theta)$;
			\ENDFOR
			\STATE Get $\bm{e}^*_{\tau}=  g([\vec{\cD_\tau},u];\theta)$;
			\STATE Get inner-task distance $d^{\text{in}}_{\tau}$ by \eqref{eq:ind};
			\STATE Get task loss $\frac{1}{m}\sum_{i=0}^{m-1} \ell(y_{\tau,i+1},g([\vec\cD_{\tau,0:i},x_{\tau,i+1}];\theta))$;
			\ENDFOR
			\STATE Get $d^{\text{in}}=\frac{1}{B}\sum_{\tau\in\bm{b}}d^{\text{in}}_\tau$ and $d^{\text{out}}$ by \eqref{eq:xd};
			\STATE Get episodic loss $\mL_e=\frac{1}{B}\sum_{\tau\in\bm{b}}\frac{1}{m}\sum_{i=0}^{m-1} \ell(y_{\tau,i+1},g([\vec\cD_{\tau,0:i},x_{\tau,i+1}];\theta))$
			\STATE Update $\theta$ by $\theta\leftarrow\theta-\nabla_\theta (\mL_e+\lambda (d^{\text{in}}-d^{\text{out}}))$.
			\ENDWHILE
		\end{algorithmic}
	\end{algorithm}

	\begin{algorithm}[ht]
		\caption{MAML w/ \TheName{}.}
		\label{alg:mamlc}
		\begin{algorithmic}
			\STATE {\bfseries Input:} Task distribution $p(\tau)$, batch size $B$, inner-task sample times $K=1$ and sampling strategy $\pi_\kappa$
			\WHILE {Not converged}
			\STATE Sample a batch of tasks $\bm{b}\sim p^B(\tau)$.
			\FOR { All $\tau\in \bm{b}$}
			\FOR {$k=1,2,\cdots,K$}
			\STATE Sample $\kappa_k$ from $\pi_\kappa(\cD^{\text{tr}}_\tau\cup\cD^{\text{val}}_\tau)$;
			\STATE Get  model representation $\bm{e}^{\kappa_k}_{\tau}=\theta-\nabla_\theta\mL(\kappa_k;h_\theta)$;
			\ENDFOR
			\STATE Get model representation $\bm{e}^*_{\tau}=\theta-\nabla_\theta\mL(\cD^{\text{tr}}_\tau\cup\cD^{\text{val}}_\tau;h_\theta)$.
			\STATE Get inner-task distance $d^{\text{in}}_{\tau}$ by \eqref{eq:ind};
			\STATE Get task-specific model $h_{\theta-\nabla_\theta\mL(\cD^{\text{tr}}_\tau;\theta)}$;
			\STATE Get validation loss $\mL(\cD^{\text{val}}_{\tau};h_{\theta-\nabla_\theta\mL(\cD^{\text{tr}}_\tau;h_\theta)})$;
			\ENDFOR
			\STATE Get $d^{\text{in}}=\frac{1}{B}\sum_{\tau\in\bm{b}}d^{\text{in}}_\tau$ and $d^{\text{out}}$ by \eqref{eq:xd};
			\STATE Get loss $\mL_{\text{\TheName}}$ by \eqref{eq:obj};
			\STATE Update $\theta$ by $\theta\leftarrow\theta-\nabla_\theta \mL$.
			\ENDWHILE
		\end{algorithmic}
	\end{algorithm}

	\begin{algorithm}[ht]
		\caption{Reptile w/ \TheName{}.}
		\label{alg:reptilec}
		\begin{algorithmic}
			\STATE {\bfseries Input:} Task distribution $p(\tau)$, batch size $B$. (inner-task sample times $K=1$ and sampling strategy $\pi_\kappa(\cD^{\text{tr}}_\tau\cup\cD^{\text{val}}_\tau)=\cD^{\text{tr}}_\tau$)
			\WHILE {Not converged}
			\STATE Sample a batch of tasks $\bm{b}\sim p^B(\tau)$.
			\FOR { All $\tau\in \bm{b}$}
			\FOR {$k=1,2,\cdots,K$}
			\STATE Sample $\kappa_k$ from $\pi_\kappa(\cD_\tau)$;
			\STATE Get  model representation $\bm{e}^{\kappa_k}_{\tau}=\theta-\nabla_\theta\mL(\kappa_k;h_\theta)$;
			\ENDFOR
			\STATE Get model representation $\bm{e}^*_{\tau}=\theta-\nabla_\theta\mL(\cD^{\text{tr}}_\tau\cup\cD^{\text{val}}_\tau;h_\theta)$.
			\STATE Get inner-task distance $d^{\text{in}}_{\tau}$ by \eqref{eq:ind};
			\ENDFOR
			\STATE Get $d^{\text{in}}=\frac{1}{B}\sum_{\tau\in\bm{b}}d^{\text{in}}_\tau$ and $d^{\text{out}}$ by \eqref{eq:xd};
			\STATE Update $\theta$ by $\theta\leftarrow\theta+\frac{1}{B}\sum_{\tau\in\bm{b}}(\bm{e}^*_{\tau}-\theta)-\lambda\nabla_\theta  (d^{\text{in}}-d^{\text{out}})$.
			\ENDWHILE
		\end{algorithmic}
	\end{algorithm}

	\begin{algorithm}[ht]
		\caption{SCNAPs w/ \TheName{}.}
		\label{alg:hyperc}
		\begin{algorithmic}
			\STATE {\bfseries Note:} Here $h_w$ corresponds to the feature extractor $f_\theta$; $H_\theta$ corresponds to the task encoder $g_\phi$ in \citep{bateni2020improved}.
			\STATE {\bfseries Input:} Task distribution $p(\tau)$, batch size $B$, inner-task sample times $K$ and sampling strategy $\pi_\kappa$.
			\STATE Pretrain $h_w$ with the mixture of all meta-training data;
			\WHILE {Not converged}
			\STATE Sample a batch of tasks $\bm{b}\sim p^B(\tau)$.
			\FOR { All $\tau\in \bm{b}$}
			\FOR {$k=1,2,\cdots,K$}
			\STATE Sample $\kappa_k$ from $\pi_\kappa(\cD^{\text{tr}}_\tau\cup\cD^{\text{val}}_\tau)$;
			\STATE Get model representation $\bm{e}^{\kappa_k}_{\tau}= H_\theta (\kappa_k)$;
			\ENDFOR
			\STATE Get model representation $\bm{e}^*_{\tau}=  H_\theta (\cD^{\text{tr}}_\tau\cup\cD^{\text{val}}_\tau)$;
			\STATE Get inner-task distance $d^{\text{in}}_{\tau}$ by \eqref{eq:ind};
			\STATE Get task-specific model by FiLM $h_{\tau}=h_{w,H_\theta (\cD^{\text{tr}}_\tau)}$;
			\STATE Get validation loss $\mL(\cD^{\text{val}}_{\tau};h_\tau)$;
			\ENDFOR
			\STATE Get $d^{\text{in}}=\frac{1}{B}\sum_{\tau\in\bm{b}}d^{\text{in}}_\tau$ and $d^{\text{out}}$ by \eqref{eq:xd};
			\STATE Get loss $\mL_{\text{\TheName}}$ by \eqref{eq:obj};
			\STATE Update $\theta$ by $\theta\leftarrow\theta-\nabla_\theta \mL$.
			\ENDWHILE
		\end{algorithmic}
	\end{algorithm}

	\begin{algorithm}[ht]
		\caption{ProtoNet w/ \TheName{} ($N$-way classification).}
		\label{alg:protoc}
		\begin{algorithmic}
			\STATE {\bfseries Input:} Task distribution $p(\tau)$, batch size $B$, inner-task sample times $K=1$ and sampling strategy $\pi_\kappa$
			\WHILE {Not converged}
			\STATE Sample a batch of tasks $\bm{b}\sim p^B(\tau)$.
			\FOR { All $\tau\in \bm{b}$}
			\FOR {$k=1,2,\cdots,K$}
			\STATE Sample $\kappa_k$ from $\pi_\kappa(\cD^{\text{tr}}_\tau\cup\cD^{\text{val}}_\tau)$;
			\STATE Calculate prototypes $\bm{c}_j=\frac{1}{|\kappa_{k,j}|}\sum_{(x_i,y_i)\in\kappa_{k,j}}f_\theta(x_i)$ for $j=1,\cdots,N$;
			\STATE Get  model representation $\bm{e}^{\kappa_k}_{\tau}=[\bm{c}_1|\bm{c}_2|\cdots|\bm{c}_N]$;
			\ENDFOR
			\STATE Calculate prototypes $\bm{c}_j=\frac{1}{|\cD_j|}\sum_{(x_i,y_i)\in\cD_j}f_\theta(x_i)$ for $j=1,\cdots,N$;
			\STATE Get model representation $\bm{e}^*_{\tau}=[\bm{c}_1|\bm{c}_2|\cdots|\bm{c}_N]$;
			\STATE Get inner-task distance $d^{\text{in}}_{\tau}$ by \eqref{eq:ind};
			\STATE Get task-specific model $h_{[\bm{c}_1|\bm{c}_2|\cdots|\bm{c}_N]}$, which gives prediction by $p(y=j\mid x)=\frac{exp(-d(f_\theta(x),\bm{c}_j))}{\sum_{j'}exp(-d(f_\theta(x),\bm{c}_{j'}))}$;
			\STATE Get validation loss $\mL(\cD^{\text{val}}_{\tau};h_{[\bm{c}_1|\bm{c}_2|\cdots|\bm{c}_N]})$;
			\ENDFOR
			\STATE Get $d^{\text{in}}=\frac{1}{B}\sum_{\tau\in\bm{b}}d^{\text{in}}_\tau$ and $d^{\text{out}}$ by \eqref{eq:xd};
			\STATE Get loss $\mL_{\text{\TheName}}$ by \eqref{eq:obj};
			\STATE Update $\theta$ by $\theta\leftarrow\theta-\nabla_\theta \mL$.
			\ENDWHILE
		\end{algorithmic}
	\end{algorithm}

	\clearpage

	\section{ICL with \TheName{}}\label{sec:icl}

	\subsection{ICL}
	ICL  is first proposed for LLMs \citep{brown2020language},
	where examples in a task are integrated into the prompt (input-output pairs) and given a new query input, the language model can generate the corresponding output. This approach allows pretrained model to address new tasks without fine-tuning the model.
	For example, given "\textit{happy->positive; sad->negative; blue->}", the model can output "\textit{negative}", while given "\textit{green->cool; yellow->warm; blue->}" the model can output "\textit{cool}".
	ICL  has the ability to learn from the prompt. Training ICL can be viewed as learning to learn, i.e., meta-learning  \citep{min2022metaicl,garg2022can,kirsch2022general}.
	More generally, the input and output are not necessarily to be natural language.
	In ICL,
	a sequence model $T_\theta$ (typically transformer~\citep{vaswani2017attention}) 
	is trained to map sequence $[x_1,y_1,x_2,y_2,\cdots,x_{m-1},y_{m-1},x_m]$ (prompt prefix) to prediction $y_m$. Given distribution $P$ of training prompt $t$, then training ICL follows an auto-regressive manner:
	\begin{align}
		\label{eq:icl-obj-origin}
		\min_{\theta}
		\mathbb{E}_{t\sim P(t)}\frac{1}{m}\sum\nolimits_{i=0}^{m-1} \ell(y_{t,i+1},T_\theta([x_{t,1},y_{t,1},\cdots,x_{t,i+1}])).\end{align}
	It has been mentioned that the training of ICL can be viewed as an instance of meta-learning  \citep{garg2022can,akyurek2023learning} as $T_\theta$ learns to learn from prompt. It has been pointed out that ICL model is meta-learner with minimal inductive bias \citep{wu2025context}.
	In this section we first formally reformulate $T_\theta$ to meta-learner $g(;\theta)$, then introduce how \TheName{} can be integrated with ICL.
	
	\vspace{-10pt}
	\subsection{A Meta-learning Reformulation}
	\vspace{-7pt}
	Denote a sequentialized $\cD$ as $\vec{\cD}$ where the sequentializer is default to bridge $p(\tau)$ and $P(t)$. Then the prompt $[x_{\tau,1},y_{\tau,1},\cdots,x_{\tau,m},y_{\tau,m}]$ can be viewed as $\vec{\cD_\tau^{tr}}$ which is providing task-specific information.
	Note that ICL does not specify an explicit output model $h(x)=g(\cD;\theta)(x)$; instead, this procedure exists only implicitly through the feeding-forward of the sequence model, i.e., task-specific prediction is given by $g([\vec\cD,x];\theta)$. 
	Thus we can reformulate the training of ICL \eqref{eq:icl-obj-origin} as:
	\begin{align}
		\label{eq:icl-obj}
		\min_{\theta}\mathbb{E}_{\tau \sim p(\tau)}
		\frac{1}{m}\sum\nolimits_{i=0}^{m-1} \ell(y_{\tau,i+1},g([\vec\cD_{\tau,0:i},x_{\tau,i+1}];\theta)).
	\end{align}
	The loss in \eqref{eq:icl-obj} can be evaluated through episodic meta-training, 
	where each task in each episode  is sampled multiple times to form $\cD^{\text{val}}_{\tau}$ and $\cD^{\text{tr}}_{\tau}$ to evaluate the episodic loss $\mL_e$ in an auto-regressive manner.
	The training of ICL thus follows the episodic meta-training (Algorithm \ref{alg:episodic-training}), 
	where the validation loss with determined $\cD^{\text{tr}}_{\tau}$ and $\cD^{\text{val}}_{\tau}$: $\mL(\cD^{\text{val}}_{\tau};g(\cD^{\text{tr}}_{\tau};\theta))$, 
	is replaced by loss validated in the auto-regressive manner: 
	$\frac{1}{m}\sum\nolimits_{i=0}^{m-1} \ell(y_{\tau,i+1},g([\vec\cD_{\tau,0:i},x_{\tau,i+1}];\theta))$.
	
	\vspace{-7pt}
	\subsection{Integrating \TheName{} with ICL}
	\vspace{-5pt}
	Since the training of ICL could be reformulated as episodic meta-training, 
	the three steps to measure \TheName{} proposed in Section \ref{sec:3step} can be also adopted for ICL, 
	but the first step to obtain model representation $\psi(g(\cD,\theta))$ needs modification.
	Due to the absence of an inner learning procedure for a predictive model for prediction $h(x)=g(\cD;\theta)(x)$, 
	representation by explicit model weights of $h$ is not feasible for ICL.
	
	To represent what $g$ learns from $\cD$, we design to incorporate $\vec{\cD}$ with a dummy input $u$, 
	which functions as a probe and its corresponding output can be readout as representation:
	\begin{align}
		\psi(g(\cD;\theta))=g([\vec\cD,u];\theta),
	\end{align}
	where $u$ is constrained to be in the same shape as $x$, and has consistent value in an episode.
	The complete algorithm of \TheName{} for ICL is  in Appendix~\ref{app:algs}.
	For example, for training a ICL model on linear regression tasks we can choose $u=\bm{1}$, and in pretraining of LLM we can choose $u=$"\textit{what is this task?}".
	From the perspective of learning to learn, \TheName{} encourages ICL to align and discriminate like it does for conventional meta-learning, 
	while the representations to evaluate inner- and inter- task distance are obtained by probing output rather than explicit model weights. 
	Thus, incorporating \TheName{} into the training process of ICL benefits the fast-adaptation and task-level generalization ability.
	From the perspective of supervised learning, 
	\TheName{} is performing unsupervised data augmentation that it introduces the dummy input and contrastive objective as additional supervision to train ICL.

	\section{Experimental Results on Synthetic Data} 
	\label{sec:exp:ana}
\begin{figure*}[ht]
		\centering
		\subfigure[Model distribution of MAML.\label{fig:distrib-maml}]{
			\includegraphics[width=0.29\textwidth]{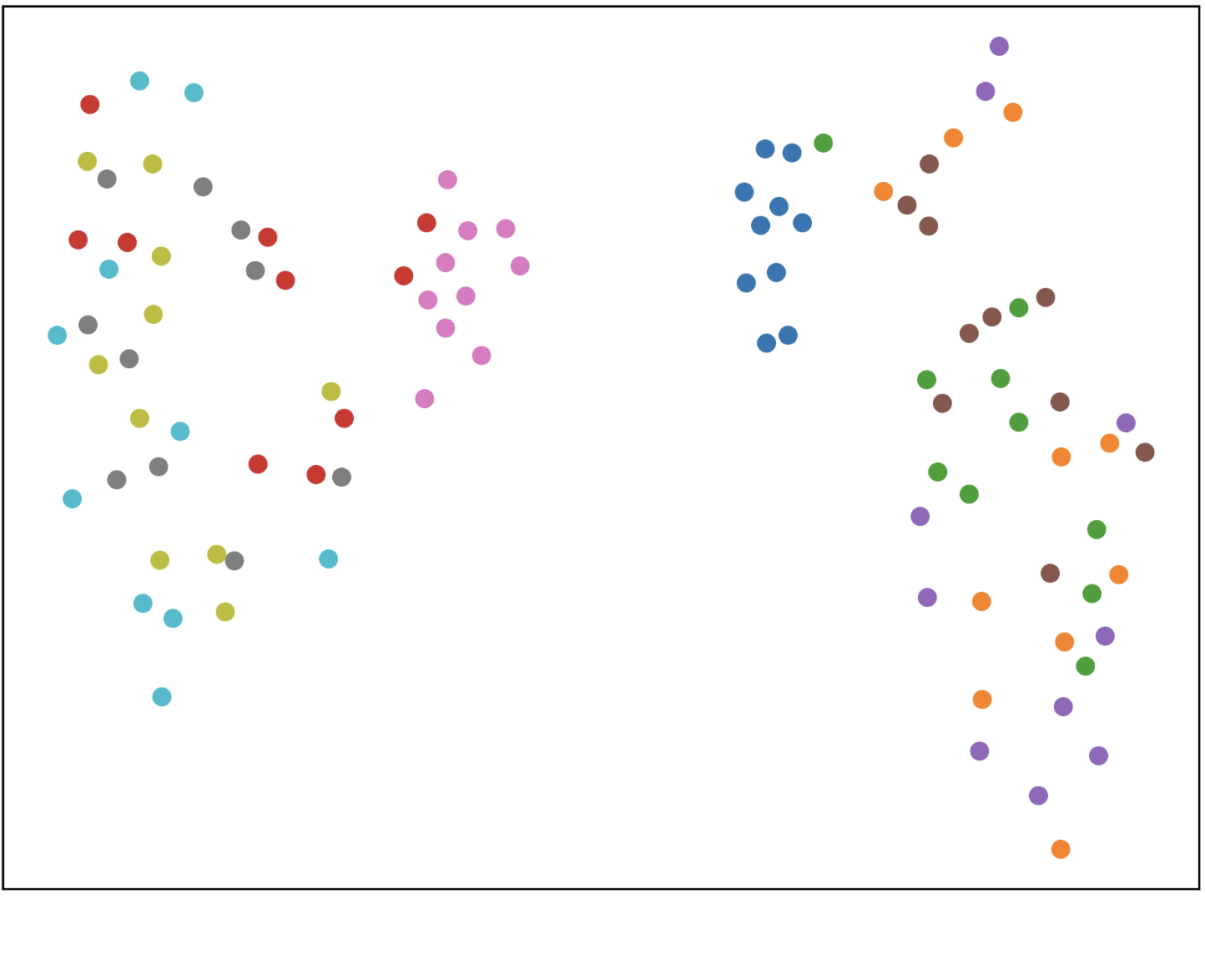}}
		\subfigure[Inner-task distance distribution.\label{fig:din}]{
			\includegraphics[width=0.32\textwidth]{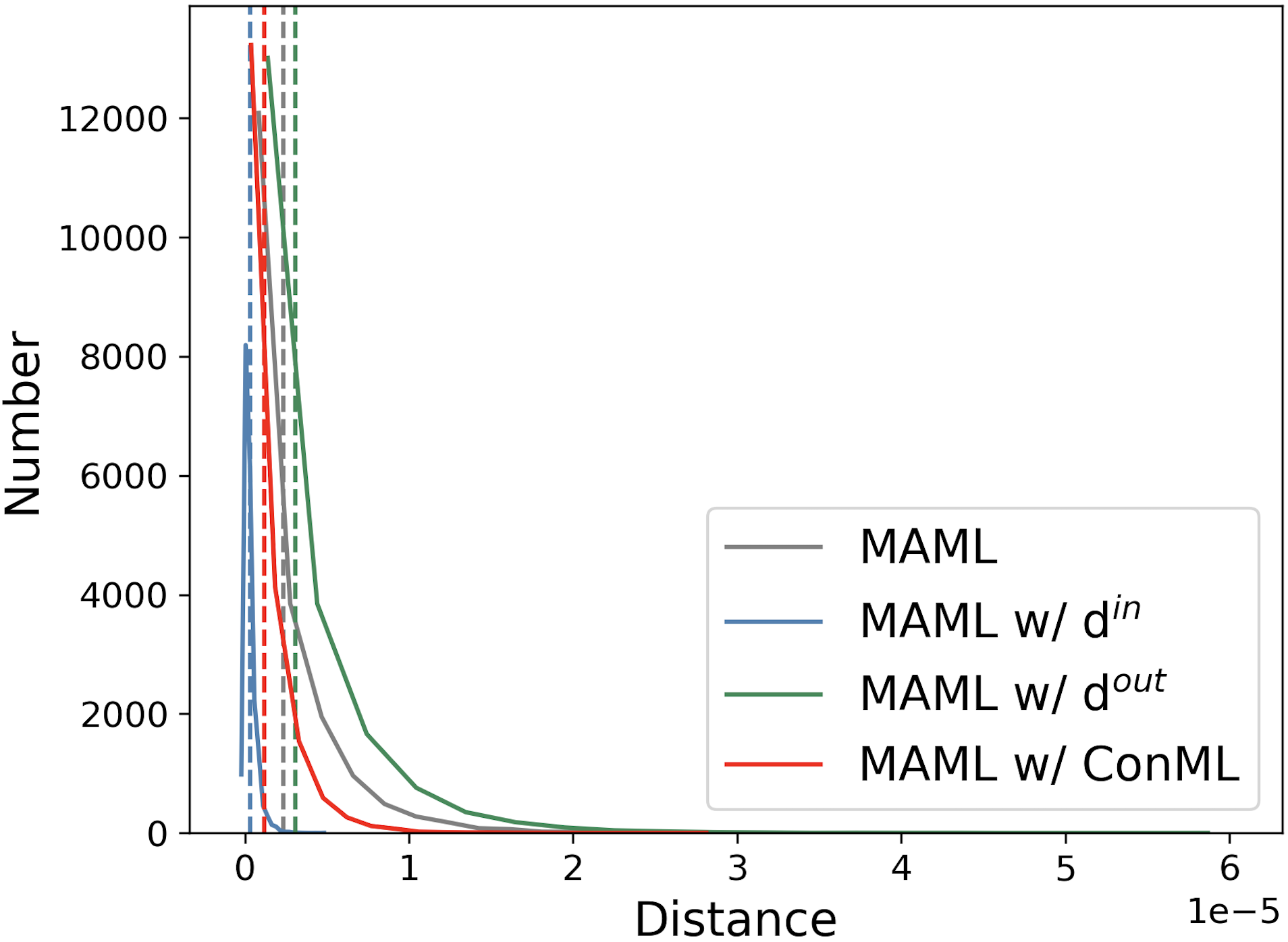}}
		\subfigure[Varying test shots.\label{fig:shot}]{
			\includegraphics[width=0.32\textwidth]{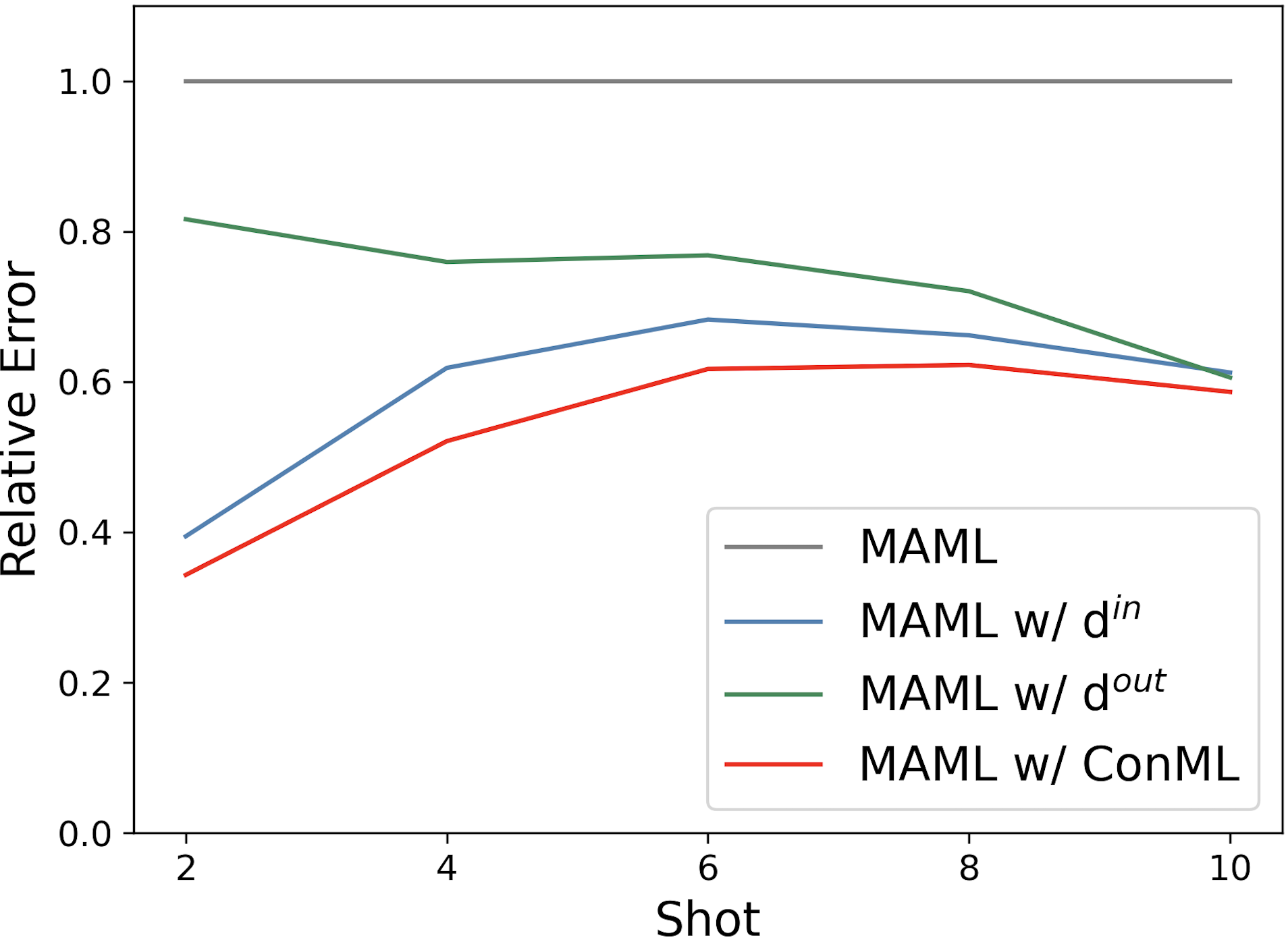}}

		\subfigure[Model distribution of MAML w/ \TheName{}.\label{fig:distrib-mamlc}]{
			\includegraphics[width=0.29\textwidth]{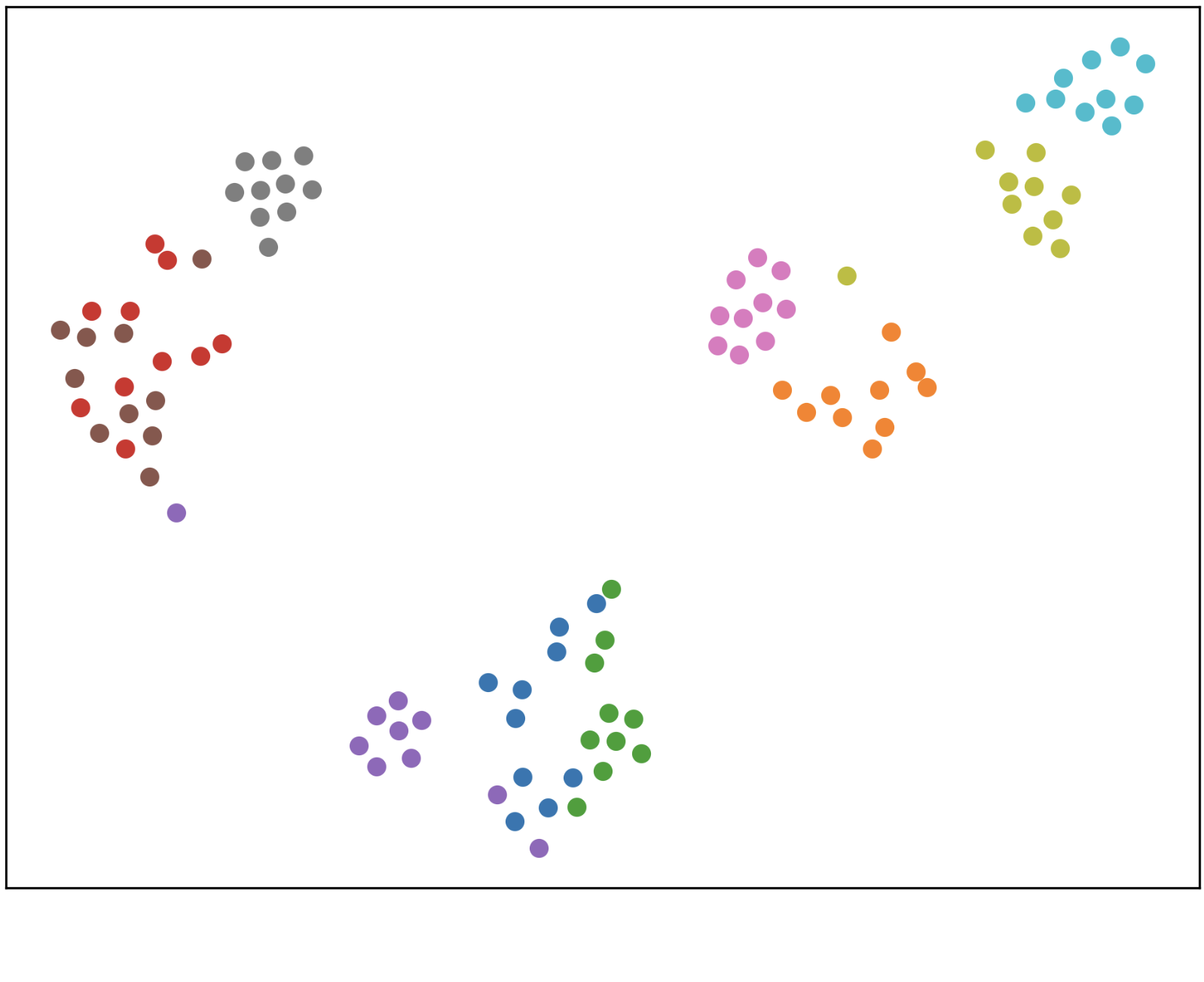}}
		\subfigure[Inter-task distance distribution.\label{fig:dout}]{
			\includegraphics[width=0.32\textwidth]{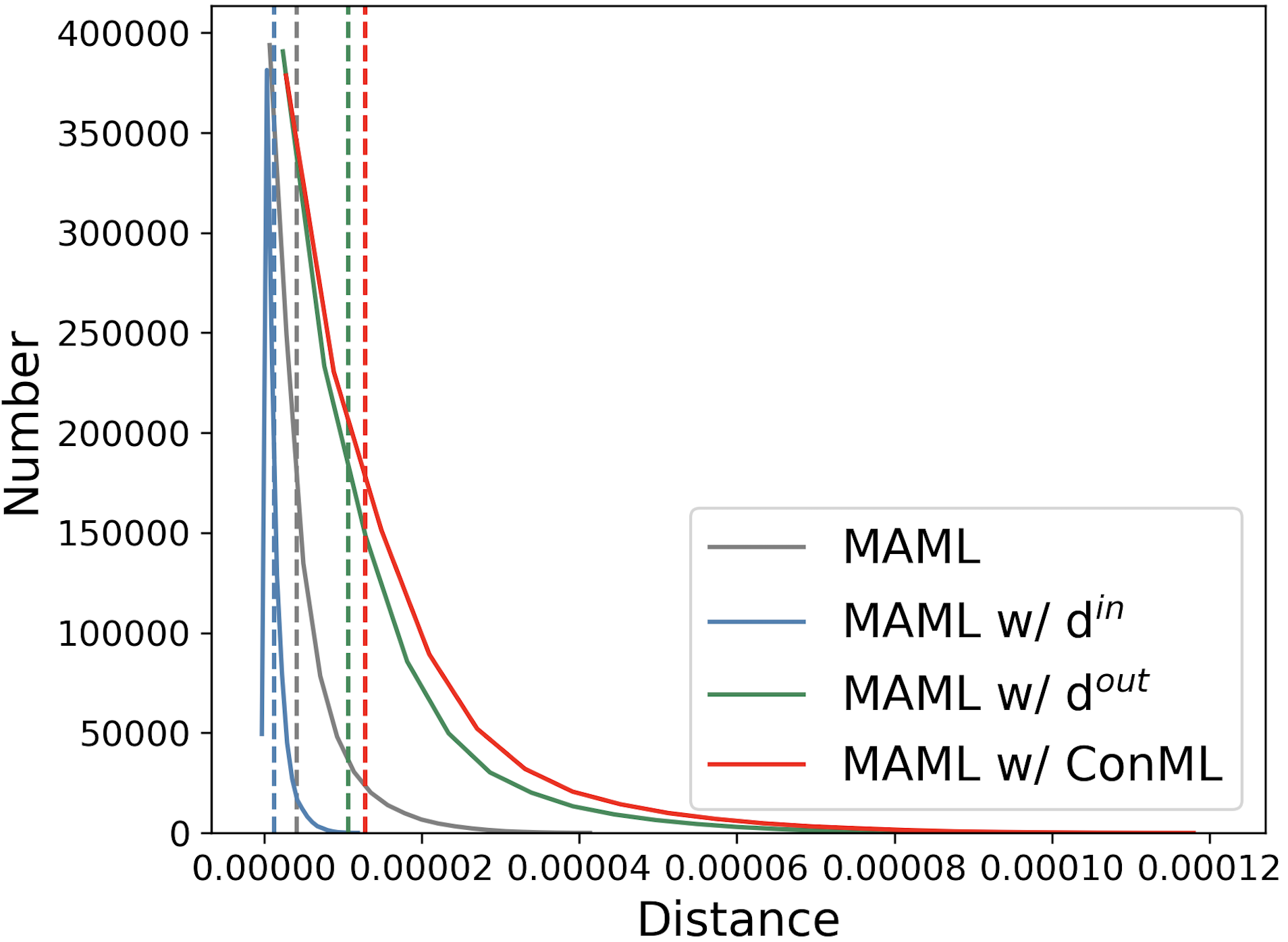}}	
		\subfigure[Varying test distribution.\label{fig:ood}]{
			\includegraphics[width=0.32\textwidth]{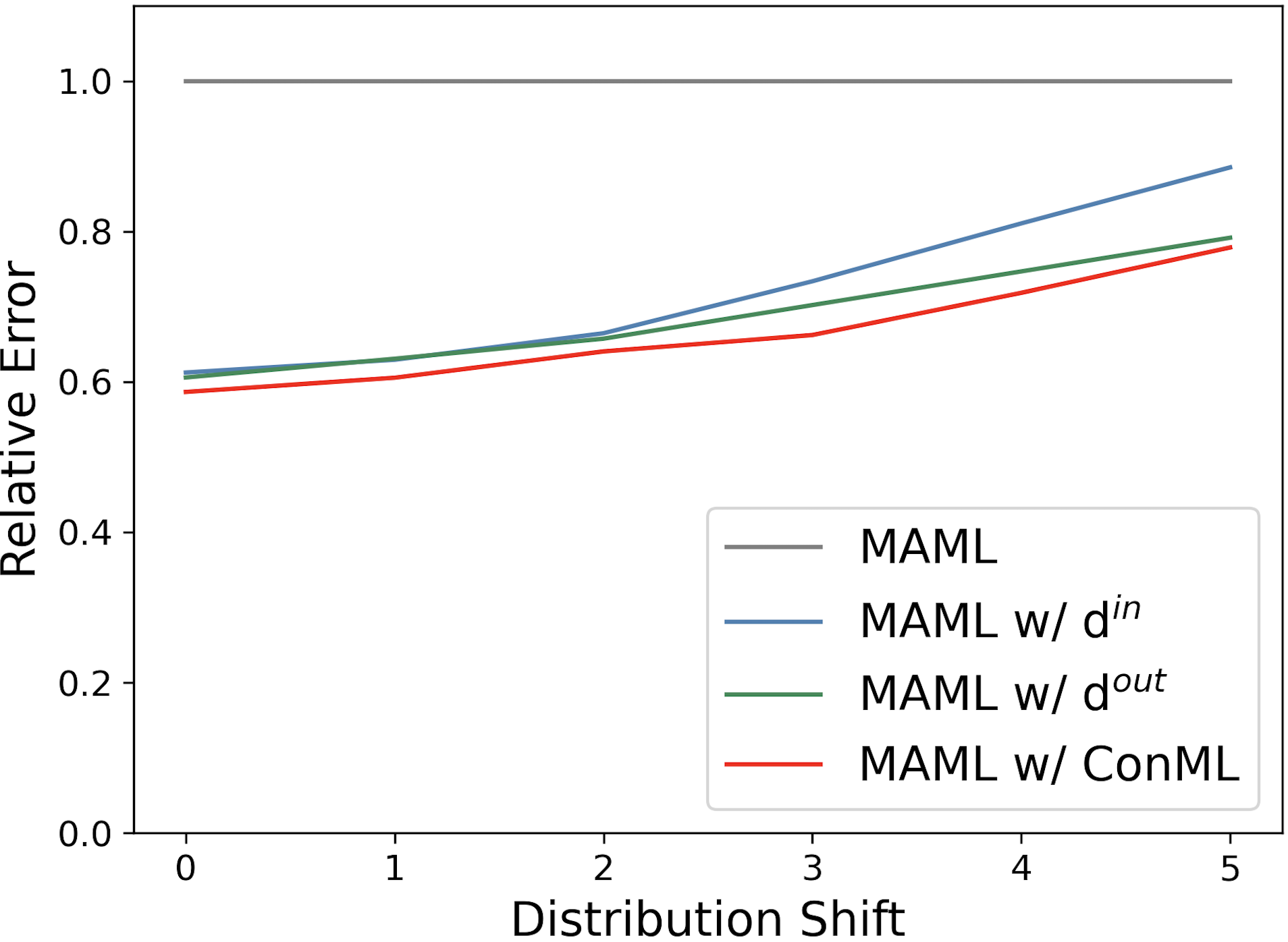}}
		\caption{Evaluation of \TheName{} on synthetic few-shot regression.}
		\label{fig:understand}
	\end{figure*}

	We begin by conducting experiments on synthetic data in a controlled setting to explain: 
	(i) Does \TheName{} enable meta-learners to develop alignment and discrimination abilities? 
	(ii) How do alignment and discrimination boost meta-learning performance?
	We take MAML w/ \TheName{} as example and investigate above questions with few-shot regression problem following the same settings in \citep{finn2017model}.
	Each task involves regressing from the input to the output of a sine wave, where the amplitude and phase of the sinusoid are varied between tasks.  The amplitude varies within $ [0.1, 5.0]$ and the phase varies within $[0, \pi]$. This synthetic regression dataset allows us to sample data and adjust the distribution as necessary for analysis. 
	The implementation of \TheName{} follows a simple intuitive setting: inner-task sampling $K=1$ and $\pi_\kappa(\cD^{\text{tr}}_\tau\cup\cD^{\text{val}}_\tau)=\cD^{\text{tr}}_\tau$, $\phi(a,b)=1-\nicefrac{a\cdot b}{\Vert a\Vert\Vert b\Vert}$ (cosine distance) and $\lambda=0.1$. 
	The meta-leaner is trained on meta-training distribution with amplitudes uniformly distributed over $[0.1, 5]$, and each training task has a fixed $N=10$. For Figure~\ref{fig:ood}, the meta-learner is tested on tasks with amplitudes uniformly distributed over
	$[0.1+\delta, 5+\delta]$, where $\delta$ is shown on the $x$-axis.

	\paragraph{Learning to learn with \TheName{} brings generalizable alignment and discrimination abilities.}
	If optimizing $d^{\text{in}}$ and $d^{\text{out}}$ does equip meta-learner with generalizable alignment and discrimination, 
	MAML w/ \TheName{} can generate more similar models from different subsets of the same task, while generating more separable models from different tasks, though these tasks are unseen.
	This can be verified by evaluating the clustering performance for model representations $\bm{e}$ from unseen tasks.
	Figure \ref{fig:distrib-maml} and \ref{fig:distrib-mamlc} visualize the distribution of these models,  where each point corresponds to the result of a subset and the same color indicates sampled from the same task. We randomly sample 10 different unseen tasks.
	For each task, we sample 10 different subsets, each containing $N=10$ samples. Using these 100 different training sets $\cD^{\text{tr}}$ as input, the meta-learner generates 100 models.
	It can be obviously observed MAML w/ \TheName{} performs better alignment and discrimination than MAML.
	To quantity the results, we also evaluate the supervised clustering performance, where task identity is used as label. Table~\ref{tab:maml-r} shows the supervised clustering performance of different metrics: Silhouette score  \citep{rousseeuw1987silhouettes} and Calinski-Harabasz index (CHI)  \citep{calinski1974dendrite}. 
	The results indicate that MAML with \TheName{} significantly outperforms standard MAML across all metrics. These findings confirm that training with \TheName{} enables meta-learners to develop alignment and discrimination abilities that generalize to meta-testing tasks. 
	
	\begin{table}[ht]
		\centering
		\caption{Meta-testing performance (MSE) on few-shot regression problem and clustering performance (Silhouette and CHI) of model representations.}
		\begin{tabular}{c||c | c||c|c}
			\toprule
			Method&	 MSE (5-shot)    & MSE (10-shot)&	Silhouette  & CHI  \\ \midrule
			MAML & $.677\pm.038$&$.068\pm.002$& $.107\pm.060$&$31.6\pm2.5$\\ \midrule
			MAML w/ \TheName{}& $\bm{.394}\pm.010$&$\bm{.040}\pm.001$&$\bm{.195}\pm.062$&$\bm{39.2}\pm2.6$\\ \bottomrule
		\end{tabular}
		\label{tab:maml-r}
	\end{table} 
	
	\paragraph{Alignment enhances fast-adaptation and discrimination enhances task-level generalizability.}
	We aim to understand the individual contributions of optimizing $d^{\text{in}}$ (alignment) and $d^{\text{out}}$ (discrimination) to meta-learning performance. In conventional unsupervised contrastive learning, both positive and negative pairs are necessary to avoid learning representations without useful information. However, in \TheName{}, the episodic loss $\mL_e$ plays a fundamental role in "learning to learn," while the contrastive objective serves as additional supervision to enhance alignment and discrimination.
	Thus, 
	we  consider 
	two variants of \TheName{}: \textbf{MAML w/ $d^{\text{in}}$} which optimizes $\mL_e$ and $d^{\text{in}}$, \textbf{MAML w/ $d^{\text{out}}$} which optimizes $\mL_e$ and $d^{\text{out}}$. 
	Figure \ref{fig:din} and \ref{fig:dout} visualize the
	distribution of $d^{\text{in}}$ and $d^{\text{out}}$ respectively, 
	where the dashed lines mark mean values. We randomly sample 1000 different unseen tasks, with 10 different subsets (each containing $N=10$ samples) per task. These subsets are aggregated into a single set of $N=100$ to obtain $\bm{e}^*_{\tau}$ for each task. 
	Smaller $d^{\text{in}}$ means better alignment and larger $d^{\text{out}}$ means better discrimination.  
	We can find that the alignment and discrimination abilities are separable, generalizable, and that \TheName{} effectively couples both.
	Figure \ref{fig:shot} shows the testing performance given different numbers of examples per task (shot).
	The results indicate that the improvement from alignment (MAML w/ $d^{\text{in}}$) is more pronounced in few-shot scenarios, highlighting its close relationship with fast-adaptation.
	Figure \ref{fig:ood} shows the out-of-distribution testing performance. 
	As the distribution gap increases, the improvement from discrimination (MAML w/ $d^{\text{out}}$) is more significant than from alignment (MAML w/ $d^{\text{in}}$), indicating that discrimination plays a critical role in task-level generalization. 
	\TheName{} leverages the benefits of both alignment and discrimination.
	
	\clearpage
	
	\section{Experimental Results Obtained Using Different Backbones}\label{app:backbone}
	In the main text, we have used the following backbones in experiment:
	\textbf{Conv4}: MAML, FOMAML, Reptile, MatchNet, ProtoNet; 
	\textbf{ResNet12}: MELR, Lastshot; 
	\textbf{ResNet18}: SCNAPs; 
	\textbf{ViT-base}: CAML.
	
	To study the effect of ConML on different backbones, we compare MAML w/o and w/ ConML, reporting the miniImageNet 5-way 1-shot accuracy.
	We specifically demonstrate the effect of equipping the model with ConML using the change in accuracy ($\Delta$ Acc). 
	\begin{table}[h]
		\centering
		\caption{Meta-testing accuracy (\%) on miniImageNet 5-way 1-shot, using different backbones.}
		\begin{tabular}{lcccc}
			\toprule
			Backbone    & Conv4   & Conv6   & ResNet12 & ResNet18 \\
			\midrule
			MAML w/o ConML    & 48.7    & 50.9    & 57.2    & 56.3    \\
			MAML w/ ConML    & 56.2    & 57.8    & 64.5    & 64.9    \\
			$\Delta$ Acc    & +7.5    & +6.9    & +7.3    & +8.6    \\
			\bottomrule
		\end{tabular}
		\label{tab:exp-backbone}
	\end{table}
	
	Table~\ref{tab:exp-backbone} shows the results.
	As the network deepens, $\Delta$ Acc shows little change. 
	Although deeper networks generally achieve higher baseline accuracy, making further improvements challenging, ConML consistently enhances performance—even outperforming shallower architectures. For instance, while ResNet18 (which may be overly deep for 1-shot MAML on miniImageNet) generalizes worse than the shallower ResNet12 without ConML, ConML boosts ResNet18's by a significant $\Delta$ Acc (+8.6\%), surpassing ResNet12 with ConML. This suggests that deeper networks can better leverage ConML's alignment and discrimination capabilities.

	\clearpage
	
	\section{Implementation Details}
	
	\subsection{Model Analysis}\label{app:hyper}
	\TheName{} optimizes the following objective: 
	$\mL_{\text{\TheName}} = \mL_e + \lambda \mL_c$,
	where \( \mL_e \) is the episodic loss,  and \( \mL_c \) is the contrastive loss.
	In the previous sections, to highlight our motivation and perform a decoupled analysis, we used the naive contrastive loss 
	$\mL_c = d^{\text{in}} - d^{\text{out}}$, with the natural cosine distance 
	$\phi(x,y) = 1 - \frac{x^\top y}{\|x\|\|y\|}$. 
	Here, we also considered a manually bounded Euclidean distance 
	$\phi(x,y) = \text{sigmoid}(\|x - y\|)$. 
	Beyond the simple contrastive loss, we incorporate the InfoNCE \citep{oord2018representation} loss for an episode with a batch $b$ containing $B$ tasks. The contrastive loss is defined as 
	$\mL_c = -\sum_{\tau \in b} \log \left( \frac{\exp(-D_{\tau}^{in})}{\exp(-D_{\tau}^{in}) + \sum_{\tau' \in b \setminus \tau} \exp(-D_{\tau,\tau'}^{out})} \right)$,
	where
	$D_{\tau,\tau'}^{out} = \phi(e_{\tau}^*, e_{\tau'}^*)$.
	In this case, we treat negative "distance" as "similarity." For the similarity metric in InfoNCE, we experiment with both cosine distance
	$\phi(x,y) = 1 - \frac{x^\top y}{\|x\|\|y\|} \text{ and Euclidean distance } \phi(x,y) = \|x - y\|$.
	
	\subsection{ICL}\label{app:exp-icl}
	We implement ICL w/ \TheName{} with $K=1$ and $\pi_\kappa([x_1,y_1,\cdots,x_n,y_n])=[x_1,y_1,\cdots,x_{\lfloor\frac{n}{2}\rfloor},y_{\lfloor\frac{n}{2}\rfloor}]$.
	To obtain the implicit representation \eqref{eq:repre-icl}, we sample $u$ from a standard normal distribution (the same with $x$'s distribution) independently in each episode. Since the output of \eqref{eq:repre-icl} is a scalar, i.e., representation ${e}\in\mathbb{R}$, we adopt distance measure $\phi(a,b)=\sigma((a-b)^2)$, where $\sigma(\cdot)$ is sigmoid function to bound the squared error. $\lambda=0.02$.
	Note that the learning of different functions (LR, DT, SLR, NN) share the same efficient and straightforward settings about \TheName{} above, which shows \TheName{} can bring ICL universal improvement with cheap implementation.  
	
	We notice that during training of LR and SLR $\lfloor\frac{n}{2}\rfloor=5$, which happens to equals to the dimension of the regression task. This means sampling by $\pi_\kappa$ would results in the minimal sufficient information to learn the task. In this case, minimizing $d^{\text{in}}$ is particularly beneficial for the fast-adaptation ability, shown as Figure \ref{fig:icl-linear} and \ref{fig:icl-slr}. This indicates that introducing prior knowledge to design the hyperparameter settings of \TheName{} could bring more advantage.

\end{document}